\documentclass[11pt]{article}

\usepackage[final]{acl}

\usepackage{times}
\usepackage{latexsym}

\usepackage[T1]{fontenc}
\usepackage[utf8]{inputenc}

\usepackage{microtype}

\usepackage{inconsolata}

\usepackage{graphicx}

\usepackage{comment}

\usepackage{lipsum}

\usepackage{float}

\usepackage{amsfonts}
\usepackage{amssymb} %
\usepackage{textcomp} %
\usepackage{verbatim}
\usepackage{pifont} %
\usepackage{fontawesome} %
\usepackage[clock]{ifsym} %
\usepackage{upgreek} %
\usepackage{xspace} %
\usepackage{upquote}
\usepackage{bbding}

\usepackage{xcolor}
\usepackage{transparent}

\usepackage[most]{tcolorbox} %

\usepackage{adjustbox}

\usepackage{xparse}
\usepackage{scalerel} %
\usepackage{graphicx}
\usepackage{pgfplots}
\usepackage{subcaption}
\usepackage{pgfplots}
\pgfplotsset{compat=1.18} %
\usepackage{pgfplotstable}
\usepgfplotslibrary{groupplots}
\usepgfplotslibrary{statistics}
\pgfplotsset{compat=newest}
\usetikzlibrary{matrix} %
\usetikzlibrary{pgfplots.fillbetween,patterns,patterns.meta} %
\usetikzlibrary{calc, positioning} %
\usepackage{tikz}
\pgfplotsset{compat=1.18,
    /pgfplots/xbar legend/.style={
    /pgfplots/legend image code/.code={%
       \draw[##1,/tikz/.cd,yshift=-0.25em]
        (0cm,0cm) rectangle (3pt,0.8em);},
   },
   /pgfplots/ybar legend/.style={
    /pgfplots/legend image code/.code={%
       \draw[##1,/tikz/.cd,yshift=-0.25em]
        (0cm,0cm) rectangle (3pt,0.8em);},
   },
}

\usepackage{array}
\usepackage{multirow}
\usepackage{colortbl}
\usepackage{arydshln} %
\usepackage{tabularx} %
\usepackage[para,online,flushleft]{threeparttable} %
\usepackage{stackengine}
\usepackage{hhline} %
\usepackage{diagbox} %
\usepackage{multicol}
\usepackage{booktabs}
\usepackage{rotating}
\usepackage{makecell}

\usepackage{bm} %
\usepackage{amsfonts} %
\usepackage{nicefrac} %

\usepackage{mathtools}

\usepackage{enumitem} %

\usepackage[nameinlink,capitalise]{cleveref}
\crefname{section}{\S\hspace{-0.2em}}{\S}

\definecolor{tablehighlight}{RGB}{239, 239, 239}

\definecolor{plotbackground}{HTML}{FBFAF9}

\definecolor{crayonRed}{HTML}{EE4266}
\definecolor{crayonBlue}{HTML}{5C95FF}
\definecolor{crayonYellow}{HTML}{FFD23F}
\definecolor{crayonGray}{HTML}{868686}

\definecolor{portscolor}{HTML}{FFAF1A}
\definecolor{replugcolor}{HTML}{FFDE70}
\definecolor{baselinecolor}{HTML}{C2E9F5}
\definecolor{darkerbaselinecolor}{HTML}{7CCFE9}

\definecolor{taclBlue}{RGB}{31,119,180}
\definecolor{taclRed}{RGB}{214,39,40}
\definecolor{taclGray}{RGB}{100,100,100}
\definecolor{taclLightBlue}{RGB}{174,199,232}
\definecolor{taclDeepBlue}{RGB}{31,119,180}

\definecolor{arcfill}{HTML}{B79482}
\definecolor{arcbackground}{HTML}{DBD8CE}
\definecolor{umlsarcfill}{HTML}{B79482}
\definecolor{umlsarcbackground}{HTML}{DBD8CE}

\definecolor{warnred}{HTML}{C0392B}

\definecolor{colSTY}{HTML}{9B59B6}            %
\definecolor{colCONSO}{HTML}{16A085}          %
\definecolor{colREL}{HTML}{D35400}            %
\definecolor{colUnfiltered}{HTML}{95A5A6}     %
\definecolor{elFont}{HTML}{0077ff}
\definecolor{elBg}{HTML}{b0f2f0}
\definecolor{srFont}{HTML}{e67e22}
\definecolor{srBg}{HTML}{fdebd0}
\definecolor{kiFont}{HTML}{d81b60}
\definecolor{kiBg}{HTML}{fce4ec}
\definecolor{condNoUMLS}{HTML}{4D4D4D}
\definecolor{condRandom}{HTML}{7A7A7A}
\definecolor{condShuffled}{HTML}{B05A7A}
\definecolor{condRelation}{HTML}{D35400}
\definecolor{condMetadata}{HTML}{1F77B4}
\definecolor{condPath}{HTML}{16A085}
\definecolor{condOracle}{HTML}{8E44AD}

\newcommand*\samethanks[1][\value{footnote}]{\footnotemark[#1]}

\newcommand{\quotes}[1]{``#1''}

\NewDocumentCommand\umlsIcon{m}{\hspace{0.8mm}\raisebox{-0.75mm}{\includegraphics[height=4mm]{umls_#1.pdf}}}

\tcbset{acronym_base/.style={
    on line,
    hbox,
    arc=2pt,
    boxrule=0pt,
    boxsep=0pt,
    left=2pt,
    right=2pt,
    top=1pt,
    bottom=1pt,
    baseline=2pt,
    fontupper=\ttfamily\bfseries
}}

\tcbset{
    el_style/.style={acronym_base, colback=elBg, colframe=elBg, fontupper=\color{elFont}\ttfamily\bfseries},
    sr_style/.style={acronym_base, colback=srBg, colframe=srBg, fontupper=\color{srFont}\ttfamily\bfseries},
    ki_style/.style={acronym_base, colback=kiBg, colframe=kiBg, fontupper=\color{kiFont}\ttfamily\bfseries}
}
\newcommand{\StageFA}[1]{{\mdseries #1}}
\newcommand{\EL}{\tcbox[el_style]{\StageFA{\faTags}\hspace{1.5pt}NEL}}
\newcommand{\SR}{\tcbox[sr_style]{\StageFA{\faSitemap}\hspace{1.5pt}KSR}}
\newcommand{\KI}{\tcbox[ki_style]{\StageFA{\faPlug}\hspace{1.5pt}KI}}

\newcommand{\UMLSring}[2]{%
  \begin{tikzpicture}[baseline=(current bounding box.center)]
    \draw[arcbackground, line width=5.5pt] (0,0) circle (.82cm);
    \ifnum#2>0
      \ifnum#2<100
        \pgfmathsetmacro{\eangle}{90 - 3.6 * #2}%
        \draw[arcfill, line width=5.5pt, line cap=butt]
          (90:.82cm) arc (90:\eangle:.82cm);
      \else  %
        \draw[arcfill, line width=5.5pt] (0,0) circle (.82cm);
      \fi
    \fi
    \node[align=center] at (0,0) {%
      \textbf{\scriptsize #1}\\[-2mm]
      \scriptsize #2\%
    };
  \end{tikzpicture}%
}

\newcommand{\ftag}[2]{%
  \begin{tikzpicture}[baseline=(T.base)]
    \node[font=\tiny\sffamily\bfseries, rounded corners=2pt,
          inner xsep=2.2pt, inner ysep=1.0pt,
          fill=#1!15, draw=#1!55, text=#1!80!black] (T) {#2};
  \end{tikzpicture}%
}

\newcommand{\best}[1]{\textbf{#1}}

\DeclareRobustCommand{\qwenfigicon}{\raisebox{-0.35em}{\includegraphics[height=1.38em]{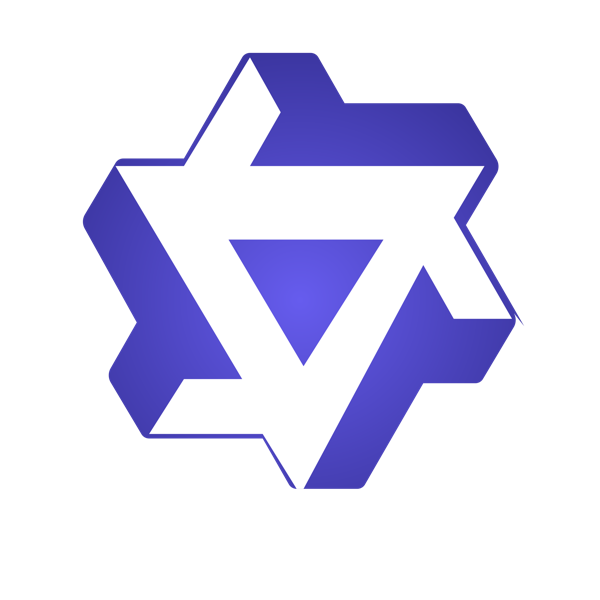}}}
\DeclareRobustCommand{\llamafigicon}{\raisebox{-0.09em}{\includegraphics[height=0.94em]{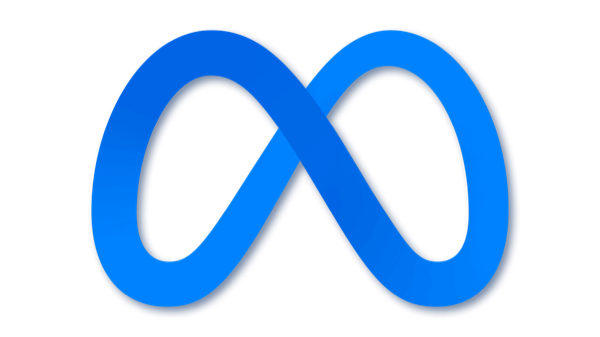}}}
\DeclareRobustCommand{\gemmafigicon}{\raisebox{-0.18em}{\includegraphics[height=1.12em]{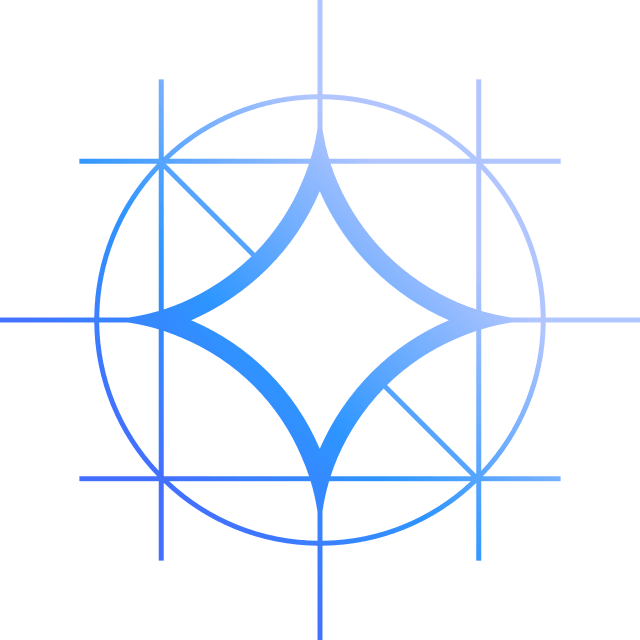}}}
\DeclareRobustCommand{\openaifigicon}{\raisebox{-0.10em}{\includegraphics[height=1.06em]{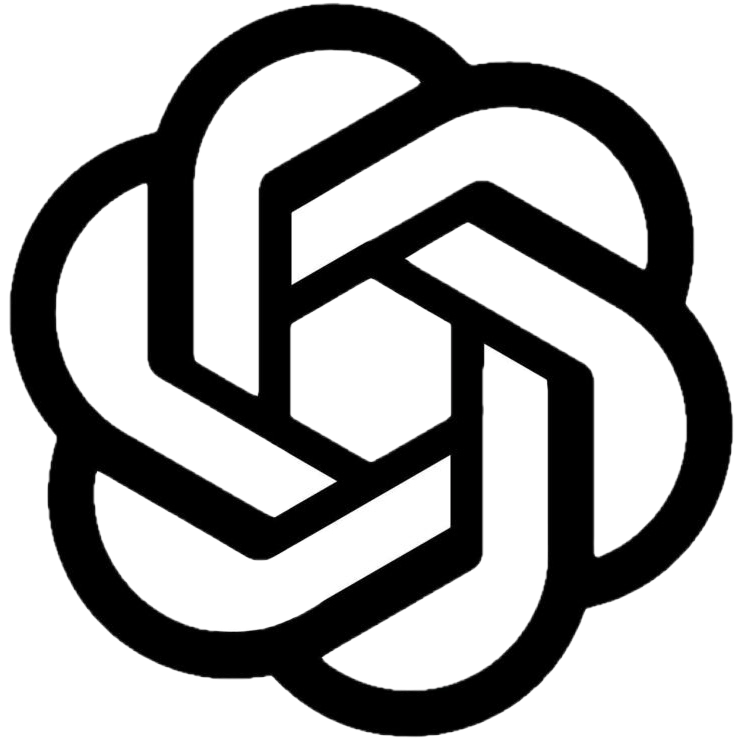}}}

\title{Are We Done with UMLS Injection into Language Models\\for Medical Question Answering?}

\author{Yiran Zeng\thanks{Equal contribution (co-first authorship).}\umlsIcon{U}\ \ \ \ Giacomo Frisoni\samethanks\umlsIcon{M}\ \ \ \ Lorenzo Molfetta\samethanks\umlsIcon{L}\ \ \ \ Gianluca Moro\samethanks\umlsIcon{S}\\
Department of Computer Science and Engineering, University of Bologna, Italy\\
\{yiran.zeng, giacomo.frisoni, lorenzo.molfetta, gianluca.moro\}@unibo.it
}

\begin{document}
\maketitle

\begin{abstract}
Maybe not.
The Unified Medical Language System (UMLS) has become an oracle for millions of clinicians and health system developers worldwide.
Researchers have increasingly used knowledge graphs derived from UMLS to enhance language models' factual accuracy and reasoning traceability in high-stakes tasks.
Medical Question Answering (MQA) is a prime testbed for integration, with benchmarks usually involving professional exercises such as disease diagnosis and drug prescription from patient history.
Despite the assumed benefits, empirical evidence shows that UMLS-based graph injection typically yields modest performance gains.
This paper reviews existing MQA work, critically analyzing the selected UMLS sources and their usage with language models, and probes each of these choices in isolation.
We uncover methodological issues never discussed in prior work, from low UMLS coverage to weak subgraph retrieval pipelines.
Based on our findings, we provide actionable recommendations to unlock the untapped potential of UMLS as an external knowledge source.\footnote{Code and literature annotations are publicly available at \href{https://disi-unibo-nlp.github.io/are-we-done-with-umls}{https://disi-unibo-nlp.github.io/are-we-done-with-uml}}
\end{abstract} %
\section{Introduction}
\label{sec:introduction}

In high-stakes medical applications, relying solely on parametric knowledge remains fundamentally problematic~\citep{DBLP:conf/acl/LabrakBMGRD24}.
Knowledge encoded in model parameters is latent, difficult to inspect, costly to update, and sensitive to pre-training data distributions~\citep{DBLP:conf/emnlp/FrisoniMMV22}.
These limitations have motivated a growing body of work on injecting external knowledge into language models (LMs), rather than expecting them to internalize all medically relevant facts~\citep{DBLP:conf/nips/LewisPPPKGKLYR020}.

Among external knowledge sources, unstructured repositories provide broad descriptive coverage but do not natively encode the relational constraints needed for controlled and inspectable reasoning~\citep{DBLP:conf/emnlp/SongYLFLYC25}.
By contrast, knowledge graphs (KGs) expose entities and typed relations explicitly, allowing evidence retrieval and multi-hop reasoning to be governed by graph structure rather than inferred from free text~\citep{DBLP:conf/emnlp/HuX0WYZCS22,DBLP:conf/iclr/SunXTW0GNSG24}.
In medicine, the Unified Medical Language System (UMLS) is the de facto structured repository for LM augmentation~\citep{DBLP:journals/nar/Bodenreider04,DBLP:journals/jbi/AfsharGGCD24}---occupying, among KG resources, the same canonical role that PubMed has among text corpora~\citep{DBLP:journals/health/GuTCLULNGP22}.

The natural testbed for UMLS-augmented LMs is medical question answering (MQA).
MQA benchmarks are often grounded in professional licensing examinations, a setting where correct answers demand not surface-level pattern matching but precise recall and compositional reasoning over clinical concepts.
Despite a rapidly growing literature claiming to leverage UMLS for MQA, reported effectiveness gains remain small, and no systematic review has examined whether they are attributable to UMLS-derived relational signal or whether they dissolve into artifacts of methodological variation.

Rather than adding one more architecture for a fractional gain, we answer that question and trace where the UMLS attribution breaks, establishing through original experiments the construction choices that would restore it.
In doing so, we reveal the skeletons in the literature's closet.
We survey closed-form and open-form MQA work (\cref{sec:reviewed_literature}).
We clarify what UMLS is (\cref{sec:what_umls_is}) and how its derivatives diverge from it (\cref{sec:what_umls_has_become}).
We then examine how UMLS data are retrieved and incorporated into LMs, exposing undisclosed issues and proposing remedies (\cref{sec:how_umls_is_used}).
We further provide experimental insights and best practices for applying UMLS in modern settings (\cref{sec:tbu}).
Our findings suggest that UMLS has not been tested under conditions where its contribution is identifiable, and that the path forward begins with acknowledging what has been obscured. %
\section{Reviewed Literature}
\label{sec:reviewed_literature}

\begin{figure}[t]
    \centering
    \includegraphics[width=\linewidth]{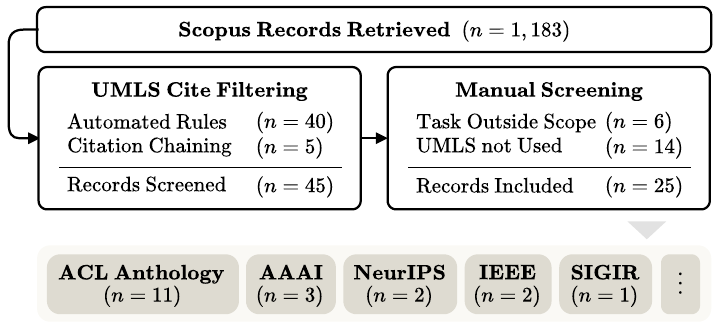}
    \caption{\textbf{Review pipeline and venue profile of the surveyed literature.} NLP-oriented conferences account for the majority of the corpus.}
    \label{fig:ic}
\end{figure}

We conduct a PRISMA-style~\citep{Pagen71} systematic review based on Scopus, as summarized in \cref{fig:ic}. 
Our search targets open-access English-language conference and journal papers published between 2020 and 2025 that use KGs or structured representations for question answering and reasoning, from which UMLS-based work is identified by citation filtering and chaining.
Papers are retained only if three conditions hold jointly: (i) UMLS plays a substantive role, (ii) the primary task is MQA or a related reasoning setting, (iii) the methodology is described in sufficient detail to support comparative analysis.
Out of the 25 included studies, 15 address closed-form settings---multiple-choice examinations such as MedQA~\citep{DBLP:journals/corr/abs-2009-13081} and PubMedQA~\citep{DBLP:conf/emnlp/JinDLCL19}---while the remaining 10 cover open-form answer generation or adjacent tasks.\footnote{Retained when their UMLS integration pipeline is methodologically equivalent to MQA usage, making design choices comparable across the stages in \cref{sec:how_umls_is_used}.}
Two medical NLP researchers independently annotated the screened papers, with 90.2\% agreement; remaining disagreements were resolved through discussion.
Reproducibility is limited: code is publicly available for only 11 studies (44\%), in line with the broader release gap documented across \textsc{*cl} venues~\citep{DBLP:conf/acl/MagnussonSD23}.
Full search strategy, eligibility criteria, and paper-level annotations are in~\cref{app:review_protocol}; the QA benchmarks these studies evaluate on are in~\cref{app:benchmark_datasets}, with examples in~\cref{tab:qa_examples}.

\section{What UMLS \textit{Really Is}}
\label{sec:what_umls_is}

First released in 1990 by the U.S.\ National Library of Medicine (NLM)~\citep{humphreys1998unified}, UMLS is a continuously updated terminology integration system whose stated mission is to help computers understand biomedical language.
It makes the constituent vocabularies jointly queryable without imposing a single unified ontology on them, so each keeps its own semantics~\citep{DBLP:journals/nar/Bodenreider04} (see \cref{app:umls_license}).
It comprises three knowledge sources---the Metathesaurus, the Semantic Network, and the SPECIALIST Lexicon---designed to be used independently or in combination.

\paragraph{Metathesaurus}
The Metathesaurus is the primary artifact.
It ships in the Rich Release Format:\footnote{Each \texttt{.RRF} file is a flat, pipe-separated text file intended to be loaded into a relational database management system.} a collection of relational files (or tables), spanning concept names, semantic types, definitions, hierarchical contexts, relationships, attributes, and source metadata.
It is updated twice a year, in May (release suffix~AA) and November (AB).
The 2025AB release, latest at analysis time, contains 23 tables covering 3.49~million unique concepts (CUIs), 17.39~million concept names, and 190~integrated source vocabularies spanning 29~languages.
Here, a \textit{concept} is a unit of meaning, whereas a \textit{concept name} denotes a lexical expression associated with that unit; a single concept routinely aggregates dozens of synonymous names from different source vocabularies.
A \textit{source vocabulary} is any external terminology, classification, coding standard, or ontology integrated into the Metathesaurus---currently ranging from SNOMED~CT and ICD-10-CM to MeSH, RxNorm, Gene Ontology, and more than 180~others.
Relations are described at two levels of granularity.
\textit{RELA} preserves the original, fine-grained predicate name asserted by the vocabulary (e.g., \texttt{has\_finding\_site}), numbering 1{,}037 distinct types in 2025AB.
\textit{REL} is NLM's own normalization: every relation, regardless of source, is mapped onto one of 13 coarse-grained categories (e.g., \texttt{is\_related\_to}).

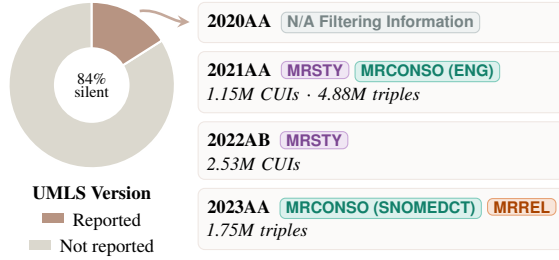
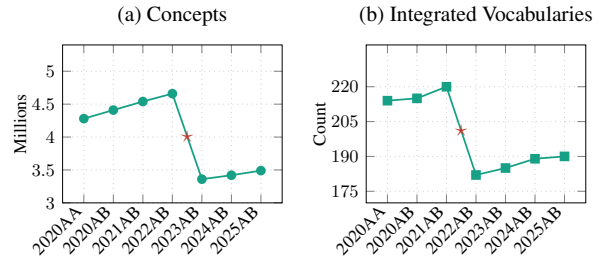
\begin{figure*}[!h]
\centering

\begin{subfigure}[t]{0.49\linewidth}
\centering
\begin{tikzpicture}[font=\small]

  \pgfmathsetmacro{\Ro}{1.1}
  \pgfmathsetmacro{\Ri}{0.5}

  \fill[umlsarcbackground, draw=white, line width=0.8pt]
    (0,0) -- (32.4:\Ro cm)
    arc[start angle=32.4, delta angle=-302.4, radius=\Ro cm]
    -- cycle;

  \fill[umlsarcfill, draw=white, line width=0.8pt]
    (0,0) -- (90:\Ro cm)
    arc[start angle=90, delta angle=-57.6, radius=\Ro cm]
    -- cycle;

  \fill[white] (0,0) circle (\Ri cm);

  \node[align=center, font=\tiny, text=black] at (0,0)
    {84\%\\[-2pt]silent};

  \node[align=center, font=\scriptsize] at (0,-1.80cm) {%
    \textbf{UMLS Version}\\[2pt]%
    \tikz[baseline=-1pt]{\fill[umlsarcfill]
      (0,0) rectangle (0.28cm,0.13cm);}\enspace
    Reported\\[2pt]%
    \tikz[baseline=-1pt]{\fill[umlsarcbackground]
      (0,0) rectangle (0.28cm,0.13cm);}\enspace
    Not reported};

  \tikzset{VB/.style={
    fill=umlsarcfill!5, draw=umlsarcfill!22, rounded corners=2.5pt,
    text width=4.6cm, align=left, font=\scriptsize, inner sep=3.5pt}}

  \node[VB, anchor=north west] (B1) at (1.4cm, 1.1cm){%
    \textbf{2020AA}\enspace\ftag{colUnfiltered}{N/A Filtering Information}};

  \node[VB, below=3pt of B1] (B2){%
    \textbf{2021AA}\enspace%
    \ftag{colSTY}{MRSTY}\;\ftag{colCONSO}{MRCONSO (ENG)}\\[1pt]%
    \textit{1.15M CUIs $\cdot$ 4.88M triples}};

  \node[VB, below=3pt of B2] (B3){%
    \textbf{2022AB}\enspace%
    \ftag{colSTY}{MRSTY}\\[1pt]%
    \textit{2.53M CUIs}};

  \node[VB, below=3pt of B3] (B4){%
    \textbf{2023AA}\enspace%
    \ftag{colCONSO}{MRCONSO (SNOMEDCT)}\;\ftag{colREL}{MRREL}\\[1pt]%
    \textit{1.75M triples}};

  \draw[->, >=stealth, umlsarcfill!80, line width=1pt,
        shorten <=3pt, shorten >=3pt]
    (61.2:\Ro cm)
    to[out=8, in=177]
    (B1.west);

\end{tikzpicture}

\subcaption{UMLS version and filtering reporting across surveyed publications. \ftag{colSTY}{MRSTY} only concepts with specified semantic types. \ftag{colCONSO}{MRCONSO ($\cdot$)} only names/synonyms with specified language or source vocabulary. \ftag{colREL}{MRREL} only specified relation types.}
\label{subfig:umls_version_papers}
\end{subfigure}%
\hfill
\begin{subfigure}[t]{0.49\linewidth}
\centering
\adjustbox{width=\linewidth}{%
\begin{tikzpicture}

\begin{groupplot}[
  group style={group size=2 by 1, horizontal sep=1.5cm},
  width=5.5cm,
  height=4.4cm,
  symbolic x coords={2020AA,2020AB,2021AB,2022AB,2023AB,2024AB,2025AB},
  xtick=data,
  x tick label style={rotate=45, anchor=east, font=\footnotesize},
  yticklabel style={font=\footnotesize},
  grid=major, grid style={dotted, gray!45},
  enlarge x limits=0.12,
  every axis plot/.style={mark size=2.0pt, line width=0.9pt},
  clip=false,
]

\nextgroupplot[
  title={(a) Concepts},
  ylabel={Millions},
  ylabel style={
    font=\footnotesize,
    at={(axis description cs:-0.13, 0.5)},
    rotate=0},
  ymin=3.0, ymax=5.40,
  ytick={3.0, 3.5, 4.0, 4.5, 5.0},
]
\addplot[mark=*, color=colCONSO] coordinates {
  (2020AA,4.28)(2020AB,4.41)(2021AB,4.54)(2022AB,4.66)
  (2023AB,3.36)(2024AB,3.42)(2025AB,3.49)
};
\path (axis cs:2022AB,4.66)
  -- node[pos=0.5, font=\large, text=warnred, inner sep=0pt] {$\star$}
  (axis cs:2023AB,3.36);

\nextgroupplot[
  title={(b) Integrated Vocabularies},
  ylabel={Count},
  ylabel style={
    font=\footnotesize,
    at={(axis description cs:-0.15, 0.5)},
    rotate=0},
  ymin=170, ymax=238,
  ytick={175, 190, 205, 220},
]
\addplot[mark=square*, color=colCONSO] coordinates {
  (2020AA,214)(2020AB,215)(2021AB,220)(2022AB,182)
  (2023AB,185)(2024AB,189)(2025AB,190)
};
\path (axis cs:2021AB,220)
  -- node[pos=0.5, font=\large, text=warnred, inner sep=0pt] {$\star$}
  (axis cs:2022AB,182);

\end{groupplot}

\end{tikzpicture}}
\subcaption{Metathesaurus content evolution (2020AA--2025AB). Drops~($\star$) mark a two-step restructuring by the National Library of Medicine: sub-source consolidation in 2022AB, followed by retirement of the associated concepts in 2023AB.}
\label{subfig:umls_version_stats}
\end{subfigure}

\caption{\textbf{UMLS data inconsistency across publications and widespread under-reporting.}
  \textit{(a)}~Reporting of UMLS release and filtering strategy.
  \textit{(b)}~Difference in CUI and vocabulary count over seven consecutive releases.}
\label{fig:umls_versioning}
\end{figure*}

\begin{figure*}[!t]
    \includegraphics[width=\linewidth]{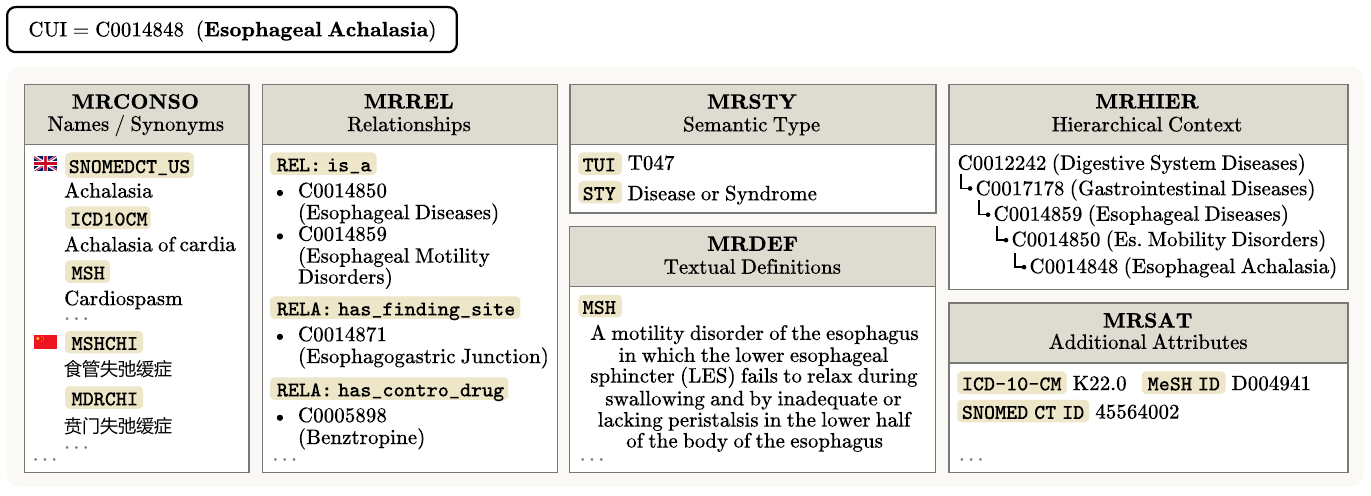}
    \vspace{6pt}
    \makebox[\linewidth][s]{\hfill
    \UMLSring{MRCONSO}{100}\hfill
    \UMLSring{MRREL}{84}\hfill
    \UMLSring{MRSTY}{20}\hfill
    \UMLSring{MRDEF}{8}\hfill
    \UMLSring{MRHIER}{0}\hfill
    \UMLSring{MRSAT}{0}\hfill
    \UMLSring{Others}{0}\hfill
    }
    \vspace{-6pt}
    \caption{\textbf{The gap between UMLS expressiveness and its use in language model injection.} \textit{Top:} the 6 CUI-centric tables of the UMLS Metathesaurus (2025AB release) illustrated on the concept Esophageal Achalasia; the remaining 17 tables are omitted as they serve infrastructure, versioning, or cross-vocabulary alignment purposes rather than encoding concept-level knowledge. \textit{Bottom:} usage frequency of each table across surveyed publications.}
    \label{fig:umls_expressiveness_coverage}
\end{figure*} 
\paragraph{Semantic Network}

The Semantic Network is an upper-level, near-static ontology of $\sim$130~semantic types (e.g., \textit{Disease or Syndrome}, \textit{Pharmacologic Substance}) connected by 54~semantic relations, organized into five non-hierarchical groupings (physically, spatially, temporally, functionally, and conceptually related)~\citep{mccray2003upper}.
While the Metathesaurus records which type has been assigned to each concept, the Semantic Network defines what that type means---its hierarchical position, its permitted relations to other types, and the ontological commitments it carries.

\paragraph{SPECIALIST Lexicon}

It is a syntactic lexicon of over one million biomedical and general English entries, encoding inflectional, derivational, and spelling-variant information for string normalization and lexical variant generation.
It underpins information extraction systems such as MetaMap~\citep{DBLP:conf/amia/Aronson01}, cTAKES~\citep{DBLP:journals/jamia/SavovaMOZSSC10}, and scispaCy~\citep{DBLP:conf/bionlp/NeumannKBA19} for mapping clinical free text to UMLS CUIs. %
\section{What UMLS \textit{Has Become}}
\label{sec:what_umls_has_become}

The reviewed literature does not employ UMLS in its original form, but relies on filtered, converted, or task-specific derivatives---yet routinely labels these artifacts simply as \quotes{UMLS,} collapsing the distinction between the source and the reduced representation.
\cref{sec:how_umls_is_used} shows how far this propagates.

\paragraph{From Databases to Knowledge Graphs}

88\% of the surveyed papers convert UMLS into a KG---a set of $(h, r, t)$ triples; the remaining 12\% query it at the concept level to retrieve definitions, semantic types, or synonyms without constructing a standalone graph~\citep{DBLP:conf/icmla/KattaRFC24,DBLP:conf/nips/WuALLSPYDMKGCM21,DBLP:conf/naacl/MichalopoulosWK21}.
Flat triple conversion has become so pervasive that it is often presented without justification, as if it were the canonical way to consume UMLS content, or the only one.

\paragraph{Untracked and Inconsistent Snapshots of an Evolving Resource}

A prerequisite for reproducibility is knowing \textit{which} UMLS a paper used; yet 84\% of the publications do not disclose a release identifier (\cref{subfig:umls_version_papers}).
The four that do cite \textit{distinct} releases: 2020AA, 2021AA, 2022AB, and 2023AA.
Releases differ substantially in content (\cref{subfig:umls_version_stats}), even when consecutive; the four adopted span $\sim$182--220 source vocabularies and $\sim$3.4--4.7M concepts.
Two systems reporting different scores may therefore have been given different knowledge, and neither paper records enough to tell which (\cref{sec:nel} and \cref{sec:injection} measure the effect).

\paragraph{Coverage Collapse}

Even setting version aside, no surveyed paper consumes the full Metathesaurus snapshot it is based on.
In practice, almost all publications draw exclusively from two of the 23 available tables---\texttt{MRCONSO} for concept names and \texttt{MRREL} for relations---leaving semantic types, definitions, hierarchical contexts, and attributes unused (\cref{fig:umls_expressiveness_coverage}).
The most common choice is to restrict \texttt{MRCONSO} to a single source vocabulary---such as SNOMED~CT or the Disease Database---abandoning the cross-source integration that is UMLS's defining contribution over any individual terminology~\citep{DBLP:conf/emnlp/Meng0CSC21,DBLP:conf/icaart/VladikaFM24}.
The cost is concrete.
For example, drug names and dosages require DrugBank or RxNorm, while pharmacological classification and anatomical site mapping require ATC; restricting to one source makes such multi-hop queries structurally impossible.
Many papers further prune the relation set to a small, manually
selected pool.
\texttt{MRREL} relation types are reduced to as few as 4 in the most restrictive case~\citep{DBLP:conf/aaaiss/DernbachAZHC24}, while the Semantic Network's types are similarly hand-picked down to as few as 2--16~\citep{DBLP:conf/www/ChoudhuryAHT23,DBLP:conf/acl-clinicalnlp/MannionSG23}, with selection criteria rarely justified in either case.
Across papers reporting filtering decisions, no two adopt the same strategy.
Consequently, author-reported graph scales diverge dramatically---CUI counts range from 1.15M to 2.53M and triple counts from 1.75M to 4.88M, a more-than-twofold spread precluding direct cross-paper comparison.
Without explicit provenance for both release and filtering decisions, observed performance variation across studies cannot be attributed to modeling choices rather than differences in the knowledge base itself. %
\section{How \quotes{UMLS} Is \textit{Actually} Used}
\label{sec:how_umls_is_used}

\begin{figure}[t]
    \centering
    \includegraphics[width=\linewidth]{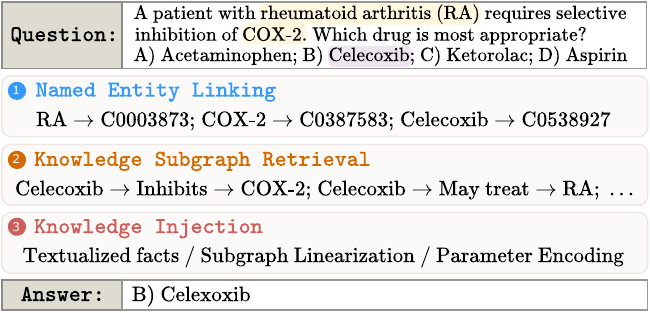}
    \caption{\textbf{Construction pattern for UMLS use in MQA.} Surveyed systems ground input text to UMLS concepts, construct a task-specific graph or evidence object, and expose the resulting representation through textual, structural, or training-time injection.}
    \label{fig:pipeline}
\end{figure}

\cref{sec:what_umls_has_become} concerns what UMLS is reduced to before any question is asked.
Once a question arrives, three stages decide which fragment of that reduced resource the model sees (\cref{fig:pipeline}).
\EL{} \textbf{Named Entity Linking} maps input spans to UMLS concepts, fixing which portions of the Metathesaurus are reachable for an instance.
\SR{} \textbf{Knowledge Subgraph Retrieval} expands those concepts into an evidence object by traversing the Metathesaurus neighborhood.
\KI{} \textbf{Knowledge Injection} determines how that object conditions the model at inference or training time.
These stages are routinely framed as engineering details, yet they decide which concepts, relations, and metadata fields constitute UMLS evidence for each prediction.
Only 6 of the 11 studies with public code (\cref{sec:reviewed_literature}) make all three stages recoverable (\cref{app:annotations_per_paper}).

\subsection{\EL{} Named Entity Linking}
\label{sec:nel}

\paragraph{Same UMLS, Different Linkers}

The first loss of UMLS evidence occurs before any graph is built.
If a question or answer option fails to link, no graph encoder, prompt format, or training objective can recover UMLS evidence for that span~\citep{DBLP:journals/tkde/ShenWH15,DBLP:conf/semweb/DubeyBCL18,DBLP:conf/bionlp/NeumannKBA19}.
Among the 17 reviewed systems with an explicit linker, scispaCy is the dominant choice ($n{=}10$), followed by string matching and neural retrieval ($n{=}2$ each), with MetaMap, cTAKES, and LLM-prompted extraction appearing once each.
None of these systems reports how much of each benchmark instance its linker covers.
Only \citet{DBLP:conf/acl-clinicalnlp/MannionSG23} evaluates linking on its target dataset, and there linking is the task itself rather than a preprocessing stage for MQA.
To fill this gap, we run scispaCy (most surveyed systems' choice) and MetaMap (the highest-coverage alternative) on the three most evaluated benchmarks (\cref{fig:question_entity_counts})---MedQA, PubMedQA, MedMCQA.
Both link against UMLS 2020AA, scispaCy's shipped default and the matching MetaMap data version.
On MedQA, MetaMap links $2\times$ more question-side CUIs and $1.5\times$ more answer-option CUIs than scispaCy; on MedMCQA, both tools link substantially fewer concepts than on MedQA, exposing an unacknowledged dataset-level variation.
PubMedQA is excluded from the answer-option analysis because its response vocabulary---\textit{yes}, \textit{no}, \textit{maybe}---does not correspond to linkable medical concepts.
On the MedQA test split, MetaMap averages 116.5 CUIs per question stem, nearly one concept per word given the benchmark's mean question length of 116.6 words~\citep{DBLP:conf/acl/FrisoniCPMM24}---a density that raises concerns about downstream noise, subgraph tractability, and context window pressure.
Linker choice alone can halve or double the UMLS evidence reachable on the same benchmark.
These counts measure reachability, not mapping quality.
We therefore rerun both linkers against UMLS 2023AA and manually label all 219 concept links they return for the 40 answer options of ten randomly drawn MedQA dev questions, marking each CUI as \textit{correct} (C) if the option expresses it in that sense, \textit{generic} (G) if expressed but clinically uninformative, and \textit{wrong} (W) otherwise.
Only 35.6\% of MetaMap's links are correct against 46.7\% for scispaCy, yet MetaMap grounds a correct concept in all 40 options against 36 for scispaCy (\cref{tab:linking_precision}).

\begin{figure}[t]\centering

\begin{tikzpicture}
\begin{axis}[
    width=.94\linewidth,
    height=4.55cm,
    xmin=-20,
    xmax=350,
    xtick={0,100,200,300},
    xlabel={Mean linked UMLS concepts (CUIs)},
    ytick={1,2,3,4,5},
    yticklabels={
        $\mathcal{A}$: MedMCQA,
        $\mathcal{A}$: MedQA,
        $\mathcal{Q}$: MedMCQA,
        $\mathcal{Q}$: PubMedQA,
        $\mathcal{Q}$: MedQA
    },
    xmajorgrids=true,
    grid=major,
    grid style={dashed, gray!45},
    axis background/.style={fill=plotbackground},
    enlarge y limits=0.16,
    major tick length=0pt,
    axis line style={draw=black!65},
    tick label style={font=\fontsize{6.4}{7.2}\selectfont},
    xlabel style={font=\fontsize{7}{8}\selectfont},
    legend style={
        draw=none,
        fill=none,
        font=\fontsize{6.6}{7.4}\selectfont,
        at={(0.5,1.02)},
        anchor=south,
        legend columns=2,
        /tikz/every even column/.append style={column sep=6pt}
    },
    error bars/x dir=both,
    error bars/x explicit,
    error bars/error bar style={thin},
    error bars/error mark options={mark size=1.4pt, thin},
    axis on top,
    clip=true
]

\draw[black!35, thin] (axis cs:-20,2.5) -- (axis cs:350,2.5);

\addplot+[
    only marks,
    mark=*,
    mark size=1.8pt,
    color=taclBlue,
    error bars/error bar style={taclBlue, thin},
    error bars/error mark options={taclBlue, mark size=1.4pt, thin}
] coordinates {
    (116.5,5.15) += (215.55,0) -= (104.45,0)  %
    (74.00,4.15)  += (60.00,0)  -= (34.00,0)   %
    (6.07,3.15)   += (70.93,0)  -= (6.07,0)    %
    (14.12,2.15)  += (60.88,0)  -= (14.12,0)   %
    (6.90,1.15)   += (34.10,0)  -= (6.90,0)    %
};
\addlegendentry{MetaMap}

\node[text=taclBlue, font=\fontsize{5.8}{6.4}\selectfont,
    anchor=west, fill=plotbackground, text opacity=1, inner sep=0.8pt]
    at (axis cs:122,5.30) {116.5};

\node[text=taclBlue, font=\fontsize{5.8}{6.4}\selectfont,
    anchor=west, fill=plotbackground, text opacity=1, inner sep=0.8pt]
    at (axis cs:79,4.30) {74.0};

\node[text=taclBlue, font=\fontsize{5.8}{6.4}\selectfont,
    anchor=west, fill=plotbackground, text opacity=1, inner sep=0.8pt]
    at (axis cs:10.5,3.33) {6.1};

\node[text=taclBlue, font=\fontsize{5.8}{6.4}\selectfont,
    anchor=west, fill=plotbackground, text opacity=1, inner sep=0.8pt]
    at (axis cs:18.5,2.33) {14.1};

\node[text=taclBlue, font=\fontsize{5.8}{6.4}\selectfont,
    anchor=west, fill=plotbackground, text opacity=1, inner sep=0.8pt]
    at (axis cs:10.0,1.33) {6.9};

\addplot+[
    only marks,
    mark=*,
    mark size=1.8pt,
    color=taclRed,
    error bars/error bar style={taclRed, thin},
    error bars/error mark options={taclRed, mark size=1.4pt, thin}
] coordinates {
    (57.67,4.85) += (111.33,0) -= (51.67,0)  %
    (69.05,3.85) += (67.95,0)  -= (35.05,0)  %
    (5.73,2.85)  += (63.27,0)  -= (5.73,0)   %
    (9.47,1.85)  += (31.53,0)  -= (9.47,0)   %
    (6.42,0.85)  += (29.58,0)  -= (6.42,0)   %
};
\addlegendentry{scispaCy}

\node[text=taclRed, font=\fontsize{5.8}{6.4}\selectfont,
    anchor=west, fill=plotbackground, text opacity=1, inner sep=0.8pt]
    at (axis cs:64,4.68) {57.7};

\node[text=taclRed, font=\fontsize{5.8}{6.4}\selectfont,
    anchor=west, fill=plotbackground, text opacity=1, inner sep=0.8pt]
    at (axis cs:74,3.68) {69.1};

\node[text=taclRed, font=\fontsize{5.8}{6.4}\selectfont,
    anchor=west, fill=plotbackground, text opacity=1, inner sep=0.8pt]
    at (axis cs:9.5,2.66) {5.7};

\node[text=taclRed, font=\fontsize{5.8}{6.4}\selectfont,
    anchor=west, fill=plotbackground, text opacity=1, inner sep=0.8pt]
    at (axis cs:11.5,1.66) {9.5};

\node[text=taclRed, font=\fontsize{5.8}{6.4}\selectfont,
    anchor=west, fill=plotbackground, text opacity=1, inner sep=0.8pt]
    at (axis cs:9.3,0.66) {6.4};

\end{axis}
\end{tikzpicture}

\caption{\textbf{Linked UMLS 2020AA concepts per instance on the test sets of three multiple-choice MQA benchmarks under two entity linkers.} Dots denote mean; bars span the minimum and maximum. $\mathcal{Q}$ and $\mathcal{A}$ denote question stems and answer options, respectively.}
\label{fig:question_entity_counts}

\end{figure}
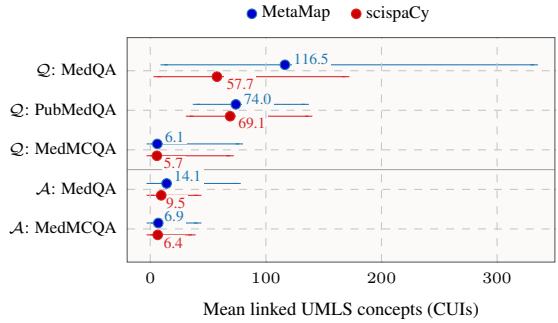 
\begin{table}[t]
\centering
\fontsize{8.5}{10}\selectfont
\begin{tabular}{@{}lrrrrrr@{}}
\toprule
\textbf{Linker} & \textbf{Links} & \textbf{C} & \textbf{G} & \textbf{W} & \textbf{C/Links} & \textbf{Grounded} \\
\midrule
MetaMap  & 180 & 64 & 45 & 71 & 35.6\% & 40/40 \\
scispaCy & 122 & 57 & 23 & 42 & 46.7\% & 36/40 \\
\midrule
Union    & 219 & 76 & 50 & 93 & 34.7\% & 40/40 \\
\bottomrule
\end{tabular}
\caption{\textbf{Manual linking precision on MedQA dev answer options (UMLS 2023AA).} 83 of the 219 links are returned by both linkers. \textit{Grounded}: options with at least one link labeled C.}
\label{tab:linking_precision}
\end{table} 

\begin{table}[!t]
  \centering
  \small
  \setlength{\tabcolsep}{4pt}
  \renewcommand{\arraystretch}{0.98}
  \resizebox{\columnwidth}{!}{%
  \begin{tabular}{lcccccc}
    \toprule
    & \multicolumn{4}{c}{\textbf{MedQA} (dev, $n{=}1{,}272$)} & \multicolumn{2}{c}{\textbf{PubMedQA} (all)} \\
    \cmidrule(lr){2-5}\cmidrule(lr){6-7}
    & \multicolumn{2}{c}{Question} & \multicolumn{2}{c}{Options} & \multicolumn{2}{c}{Question} \\
    \cmidrule(lr){2-3}\cmidrule(lr){4-5}\cmidrule(lr){6-7}
    \textbf{Configuration} & $\mu$ & distinct & $\mu$ & distinct & $\mu$ & distinct \\
    \midrule
    \multicolumn{7}{l}{\textit{UMLS release} (all sources)} \\
    2020AA$^\dagger$ & 57.67 & 9{,}338  & 9.96  & 6{,}913 & 68.46 & 13{,}536 \\
    2021AA           & 71.76 & 11{,}760 & 12.46 & 8{,}794 & 78.77 & 16{,}453 \\
    2022AB           & 67.18 & 11{,}232 & 11.98 & 8{,}487 & 75.54 & 15{,}851 \\
    2023AA           & 67.53 & 11{,}372 & 12.13 & 8{,}576 & 76.46 & 16{,}047 \\
    \midrule
    \multicolumn{7}{l}{\textit{Source subset} (2023AA)} \\
    All sources      & 67.53 & 11{,}372 & 12.13 & 8{,}576 & 76.46 & 16{,}047 \\
    Level-0          & 62.63 & 9{,}940  & 10.97 & 7{,}562 & 72.13 & 14{,}246 \\
    DrugBank+DDB     & 0.99  & 321      & 0.77  & 474     & 0.96  & 411      \\
    \bottomrule
  \end{tabular}%
  }
  \caption{\textbf{Linked UMLS concepts per instance across four releases and two source subsets.} scispaCy with the mention model held fixed and only the concept knowledge base swapped $\dagger$: its shipped default). $\mu$: mean per instance, with answer-side counts taken as the union across the four options. PubMedQA contributes question-side coverage only.}
  \label{tab:release_source_coverage}
\end{table} 

\paragraph{Same Linker, Different UMLS}

The same linker also links differently when the UMLS release or source subset changes.
Measuring this requires a linker whose mention detection is separate from its concept lookup; scispaCy, the most common linker in the reviewed systems, qualifies.
We keep its span model fixed and swap only the knowledge base, so any change in the linked concepts comes from UMLS alone (\cref{tab:release_source_coverage}).
Moving from the 2020AA default to 2021AA raises the mean linked concepts per question from 57.7 to 71.8 on MedQA and from 68.5 to 78.8 on PubMedQA, and each later release changes the linked set for at least 99.4\% of MedQA questions and every PubMedQA question.
Source restriction can cut far deeper.
Keeping only \mbox{Level-0} sources, which need no license beyond UMLS, already loses 6--7\% of the linked concepts.
\mbox{QA-GNN}~\citep{DBLP:conf/naacl/YasunagaRBLL21}, GREASELM~\citep{DBLP:conf/iclr/0001BYRLML22}, and \citet{DBLP:conf/www/0006D0Y24} keep only \mbox{DrugBank} and the Disease Database (DDB), a choice that leaves about one linked concept per question, a loss above 98\%.

\paragraph{Question-Side Versus Answer-Side Grounding}

Multiple-choice MQA amplifies the coverage gap through an informational asymmetry.
Question stems are multi-sentence clinical vignettes with dense lexical context; answer options are brief fragments, lacking the context linkers need to resolve ambiguous spans~\citep{DBLP:journals/corr/abs-2009-13081}.
We inspected the released preprocessing artifacts of QA-GNN~\citep{DBLP:conf/naacl/YasunagaRBLL21}, GREASELM~\citep{DBLP:conf/iclr/0001BYRLML22}, and DRAGON~\citep{DBLP:conf/nips/YasunagaBR0MLL22}---three systems built on reasoning paths \textit{from} question concepts \textit{to} answer concepts, all using scispaCy.
On the MedQA test split, QA-GNN links \textit{no} entity in 3.3\% of question stems, against 58.8\% of answer options.
GREASELM shows nearly the same rates, while DRAGON, which links against a broader source vocabulary, lowers the answer-side rate to 24.5\%.
For most instances, no question-to-answer path can be constructed, because the answer side never entered the UMLS space.
The architectural premise thereby fails silently on the majority of the benchmark.
In our runs with MetaMap in place of scispaCy, only ${\sim}$1\% of answer options remain unlinked in every split (\cref{app:appendix_linking_coverage}).

\subsection{\SR{} Knowledge Subgraph Retrieval}
\label{sec:subgraph}

\paragraph{Expansion as Evidence Construction}

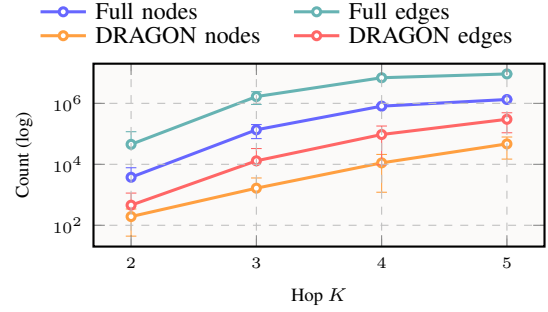
\begin{figure}[t]
\centering
\begin{subfigure}[t]{\linewidth}
    \centering
    \begin{tikzpicture}
        \footnotesize
        \node[draw=none, inner sep=2pt, align=center, text width=.92\linewidth] {
        \begin{tabular}{@{}ll@{}}
        \ref{plot:khop_full_nodes} Full nodes &
        \ref{plot:khop_full_edges} Full edges \\[-1pt]
        \ref{plot:khop_dragon_nodes} DRAGON nodes &
        \ref{plot:khop_dragon_edges} DRAGON edges
        \end{tabular}
        };
    \end{tikzpicture}
\end{subfigure}\\[0.5mm]
\begin{tikzpicture}
\begin{semilogyaxis}[
    width=0.98\columnwidth,
    height=4cm,
    xlabel={Hop $K$},
    ylabel={Count (log)},
    xtick={2,3,4,5},
    ymin=20,
    ymax=20000000,
    clip=true,
    clip mode=individual,
    grid=both,
    grid style=dashed,
    axis on top=true,
    axis background/.style={fill=plotbackground},
    thick,
    ylabel style={font=\fontsize{7}{8}\selectfont},
    xlabel style={font=\fontsize{7}{8}\selectfont},
    tick label style={font=\fontsize{6}{7}\selectfont},
    error bars/y dir=both,
    error bars/y explicit,
    error bars/error bar style={thin},
    error bars/error mark options={
        rotate=90, mark size=2pt, thin
    },
]

\addplot[color=blue!60, line width=1.15pt, mark=*, mark size=1.5pt, mark options={fill=white}] coordinates {
    (2,  3755)    +- (0, 3966)
    (3,  135535)  +- (0, 65810)
    (4,  809475)  +- (0, 103317)
    (5,  1336309) +- (0, 34770)
};
\label{plot:khop_full_nodes}

\addplot[color=teal!60, line width=1.15pt, mark=*, mark size=1.5pt, mark options={fill=white}] coordinates {
    (2,  45153)   +- (0, 72225)
    (3,  1666105) +- (0, 744238)
    (4,  6917457) +- (0, 530624)
    (5,  9221891) +- (0, 116972)
};
\label{plot:khop_full_edges}

\addplot[color=orange!75, line width=1.15pt, mark=*, mark size=1.5pt, mark options={fill=white}] coordinates {
    (2,  193)   +- (0, 149)
    (3,  1647)  +- (0, 1915)
    (4,  11145) +- (0, 9933)
    (5,  46719) +- (0, 31905)
};
\label{plot:khop_dragon_nodes}

\addplot[color=red!60, line width=1.15pt, mark=*, mark size=1.5pt, mark options={fill=white}] coordinates {
    (2,  456)    +- (0, 681)
    (3,  13000)  +- (0, 19913)
    (4,  94897)  +- (0, 85237)
    (5,  299594) +- (0, 191151)
};
\label{plot:khop_dragon_edges}

\end{semilogyaxis}
\end{tikzpicture}
\caption{\textbf{Growth of UMLS neighborhoods under $K$-hop expansion.} Mean node and edge counts over PubMedQA test instances, seeded by scispaCy, for full UMLS 2025AB and the DRAGON-filtered subset. Error bars denote mean $\pm$ one standard deviation. Log scale.}
\label{fig:khop_scale}
\end{figure}

Entity linking returns a set of disjoint seed CUIs.
Expansion connects them through the Metathesaurus: from each seed, it follows \texttt{MRREL} relations for up to $K$ steps, collecting every node and edge along the way.
Nodes outside the seed set but lying on a path between two seeds are bridge nodes: at $K{=}2$, each two-hop path passes through exactly one.
Seeds, bridges, and their connecting edges form the \textit{evidence subgraph} passed to the model.
The term \textit{retrieval} refers to the work involved in selectively drawing knowledge from UMLS into instance-specific structured contexts.
A prior decision determines which concepts are eligible as seeds.
Three patterns recur.
The most common pools question and answer CUIs into a single seed set and retains all bridge nodes reachable within $K$ hops between \textit{any} pair of linked entities~\citep{DBLP:conf/naacl/YasunagaRBLL21,
DBLP:conf/iclr/0001BYRLML22,DBLP:conf/emnlp/ZhangD0H22,
DBLP:conf/aaai/0001SWWHWCX24}.
A more restrictive alternative encodes multi-hop paths only when they link a question concept to an answer concept~\citep{DBLP:conf/aaai/ParkCKPKJKK23}.
In open-ended settings or benchmarks such as PubMedQA, where no answer-option concepts exist, the expansion necessarily bridges among question concept pairs only.
Most systems adopt $K{=}2$, a convention inherited from earlier graph-reasoning work on commonsense KGs~\citep{DBLP:conf/emnlp/FengCLWYR20,DBLP:conf/aaai/SpeerCH17} that are orders of magnitude sparser than UMLS.
Raising $K$ improves the odds of capturing clinically relevant multi-hop connections, but inflates subgraph size exponentially, introducing noise and rendering graph encoding computationally intractable.
To quantify this growth, we measured evidence subgraph size across $K$ on PubMedQA, with seeds from scispaCy under a fixed UMLS release (\cref{fig:khop_scale}).
DRAGON is the only system releasing its filtered Metathesaurus snapshot with a documented configuration, so its KG is the sole UMLS derivative that can be compared against the full 2025AB.
Under full UMLS 2025AB, the subgraph grows from thousands of nodes at $K{=}2$ to more than one million at $K{=}5$; under the DRAGON-filtered subset, it still expands from hundreds to tens of thousands, confirming that the explosion is a structural property of UMLS density.
In multiple-choice MQA, the connectivity rule introduces a further effect: at fixed $K{=}2$ on MedQA, pooling all linked concepts as seeds produces $87{\times}$ more nodes than restricting expansion to question-to-answer paths on full UMLS, and $60{\times}$ more on the DRAGON-filtered subset (\cref{tab:connectivity}).
The expanded neighborhood is then pruned: QA-GNN, GREASELM, and DRAGON score bridge nodes by relevance to the context and truncate the subgraph to a fixed budget of 200 nodes (\cref{app:pruning_details}).
The downstream impact is directly quantifiable from their released preprocessing material.
The pre-pruning graph contains approximately 298 nodes per question-option pair on average; the 200-node budget is exceeded in 70--73\% of evaluation instances.
Most critically, only ${\sim}1.5$ answer-side concepts survive per pair after pruning---a figure that compounds the answer-side linking failures documented in \cref{sec:nel}.
Option-side evidence is thus filtered twice before inference, first by entity linking and then by node budget truncation, leaving the discriminating side of every multiple-choice question structurally impoverished at both stages.

\begin{table}[t]
  \centering
  \small
  \setlength{\tabcolsep}{5pt}
  \renewcommand{\arraystretch}{1.15}
  \resizebox{0.48\textwidth}{!}{
  \begin{tabular}{llrr}
    \toprule
    \textbf{KG} & \textbf{Connectivity rule}
      & \textbf{Nodes} ($\mu \pm \sigma$)
      & \textbf{Edges} ($\mu \pm \sigma$) \\
    \midrule
    \multirow{3}{*}{\shortstack[l]{UMLS\\2025AB}}
      & All linked concepts        & $46{,}871 \pm 24{,}465$ & $385{,}342 \pm 268{,}713$ \\
      & Question$\to$answer only   & $537 \pm 1{,}007$       & $4{,}981 \pm 16{,}715$    \\
      & \textit{Ratio}
                                 & \textbf{\textit{87$\times$}} & \textbf{\textit{77$\times$}} \\
    \cmidrule(lr){1-4}
    \multirow{3}{*}{DRAGON}
      & All linked concepts        & $5{,}750 \pm 3{,}197$   & $16{,}823 \pm 9{,}444$ \\
      & Question$\to$answer only   & $96 \pm 58$             & $94 \pm 143$ \\
      & \textit{Ratio}
                                 & \textbf{\textit{60$\times$}} & \textbf{\textit{179$\times$}} \\
    \bottomrule
  \end{tabular}
  }
  \caption{\textbf{MedQA subgraph scale under two connectivity rules.} Node and edge counts at $K{=}2$ under pooled all-linked-concept versus question-to-answer retrieval.}
  \label{tab:connectivity}
\end{table}
 
\paragraph{Evidence Without Grounding}

The failures documented in \cref{sec:nel} have a direct consequence for subgraph construction: when question-to-answer path expansion is attempted but one or both sides carry no linked concepts---a condition affecting the majority of answer-option sets---systems must decide what to pass to the model in place of UMLS evidence.
This decision is never disclosed in any reviewed paper; it surfaces only through inspection of released code.
The observed fallbacks are arbitrary: systems insert dummy nodes as placeholder 
seeds~\citep{DBLP:conf/naacl/YasunagaRBLL21,DBLP:conf/iclr/0001BYRLML22,DBLP:conf/emnlp/ZhangD0H22,DBLP:conf/nips/YasunagaBR0MLL22} using a fixed concept such as \textit{Serine} when no question entity is found, and \textit{Ethanol} when no answer entity survives---neither of which corresponds to a high-degree hub or any principled selection criterion we could identify.
The resulting subgraph bears no semantic connection to the question-answer pair under evaluation.
We compare QA-GNN on MedQA with two alternative fallback policies over three seeds: test accuracy is 38.0$\pm$0.5\% with a random CUI placeholder and 37.9$\pm$0.7\% with no retrieval, against 37.4$\pm$0.7\% with the original \textit{Serine}/\textit{Ethanol} placeholders.
Changing the placeholder or skipping retrieval therefore yields similar mean accuracy, even though none of these fallbacks carries evidence related to the question.

\subsection{\KI{} Knowledge Injection}
\label{sec:injection}

\paragraph{Injection Is Not One Operation}

Knowledge injection is the most visible UMLS label in benchmark tables, but it is also the stage least able to compensate for upstream evidence loss.
By the time UMLS reaches a model, linking, retrieval, pruning, and fallback rules have already decided which concepts, relations, and metadata fields are present.
We distinguish three uses of UMLS.
\textit{Textual linearization} serializes the evidence object into the prompt, so the reader sees UMLS as tokens.
\textit{Structural injection} retains it as a graph, present at inference and encoded by a dedicated module.
\textit{Parameter encoding} consumes UMLS once, offline, as a training signal; at inference time, the knowledge is whatever survived in the weights.
Among systems with explicit injection mechanisms, our taxonomy counts 5 linearization systems, 9 structural-injection systems, and 7 parameter-encoding systems (\cref{app:injection_taxonomy}).
A single \quotes{UMLS injection} label conflates the evidence object, the integration mechanism, and the claim being made about the resource, leaving benchmark gains as statements about whole pipelines.

\paragraph{What Each Paradigm Tests}

\textit{Textual linearization} evaluates UMLS content together with the scheme expressing it.
Performance depends on which tables are verbalized and how much structure survives the rewrite~\citep{DBLP:conf/nips/LewisPPPKGKLYR020,DBLP:conf/iclr/FatemiHP24,DBLP:journals/jbi/CuiLXWMYYKLZQHFMHWY25,DBLP:conf/coling/ZafarME25}.
Most linearized records preserve \texttt{MRCONSO} names and \texttt{MRREL} relations, omitting definitions, semantic types, hierarchical context, attributes, and provenance.
The evidence view is consequently part of the claim: it determines whether the model receives UMLS as a sparse relation list or as a richer concept-centered record.
\textit{Structural injection} comes closest to evaluating UMLS as a relational resource.
Systems in this class differ in where the evidence subgraph meets the text: some encode the subgraph into a vector that conditions the LM~\citep{DBLP:conf/coling/YadavPS20,laube-eliasmith-2024-qavsa,DBLP:conf/bibm/SenguptaHCSM24}, some reason over the LM output together with the graph in a separate module~\citep{DBLP:conf/naacl/YasunagaRBLL21,DBLP:conf/emnlp/ZhangD0H22,DBLP:conf/aaai/ParkCKPKJKK23}, and some fuse graph and text representations across LM layers~\citep{DBLP:conf/iclr/0001BYRLML22,DBLP:conf/nips/YasunagaBR0MLL22}.
Parameter encoding asks a different question again, consistent with adapter-based knowledge-enhanced LM formulations~\citep{DBLP:conf/ic3k/FichtlVG24}: whether UMLS-derived supervision can be internalized at training time~\citep{DBLP:conf/icaart/VladikaFM24,DBLP:conf/emnlp/WangBWYLCC24,DBLP:conf/acl-clinicalnlp/MannionSG23}.
It also inherits a versioning problem, since knowledge absorbed into weights does not track later UMLS releases~\citep{DBLP:conf/emnlp/TangFZ19,DBLP:journals/corr/abs-2503-05212}.

\paragraph{Retraining Systems Under Seven UMLS Constructions}
Everything so far characterizes the evidence object and how it reaches the model.
None of it touches the number a paper reports.
Whether UMLS construction choices move a benchmark score is a separate question.
We select QA-GNN~\citep{DBLP:conf/naacl/YasunagaRBLL21} as the open-source reference implementation of the LM+KG paradigm for MQA, the baseline against which \citet{DBLP:conf/iclr/0001BYRLML22, DBLP:conf/emnlp/ZhangD0H22, DBLP:conf/aaai/ParkCKPKJKK23} report their gains, and GREASELM~\citep{DBLP:conf/iclr/0001BYRLML22}, which replaces its late fusion with layer-wise LM--graph interaction.
We retrain both end to end on MedQA under seven UMLS constructions over three random seeds, on the SapBERT encoder~\citep{DBLP:conf/naacl/LiuSMBC21} they share, with architecture, hyperparameters, and schedule at published values and each NEL paired with a graph from the same release (\cref{tab:endtoend}).
Our 2020AA QA-GNN run reproduces the published 38.0, and accuracy does not separate thereafter: test accuracy spans 37.5 to 38.7 across four releases and two source subsets, while an empty subgraph retaining only the two placeholder concepts of \cref{sec:subgraph} reaches 37.4, within seed variance of every configuration we tried.
GREASELM shows the same insensitivity to construction, spanning 37.7 to 39.3 across the six graph configurations, but its empty subgraph falls below all of them, so whether a graph helps at all is system-dependent.
Set against the published record, QA-GNN gains 0.8 points over its SapBERT baseline and GREASELM 0.5 over QA-GNN, so the increments grounding this line of work are no larger than the variation from a download choice.
We measure this on two GNN-era systems under structural injection; \cref{sec:tbu} investigates LLM readers under textual linearization.

\begin{table}[t]
\centering
\small
\setlength{\tabcolsep}{3pt}
\resizebox{\columnwidth}{!}{
\begin{tabular}{llccccccc}
\toprule[\lightrulewidth]
& & \multicolumn{4}{c}{\textbf{Release}} & \multicolumn{2}{c}{\textbf{Subset}} & \textbf{No Graph} \\
\cmidrule(lr){3-6}\cmidrule(lr){7-8}\cmidrule(lr){9-9}
& & 20AA & 21AA & 22AB & 23AA & Lv-0 & \makecell{DrugBank\\+DDB} & Empty \\
\midrule
\multirow{2}{*}{QA-GNN} & Dev  & 38.8\tiny{$\pm$.3} & 38.2\tiny{$\pm$.8} & 38.4\tiny{$\pm$1.0} & 38.0\tiny{$\pm$.7} & 37.8\tiny{$\pm$.6} & 38.3\tiny{$\pm$.3} & 38.9\tiny{$\pm$.6} \\
                        & Test & 38.1\tiny{$\pm$.4} & 37.6\tiny{$\pm$.6} & 38.7\tiny{$\pm$.5} & 38.0\tiny{$\pm$.9} & 37.5\tiny{$\pm$1.0} & 38.5\tiny{$\pm$.6} & 37.4\tiny{$\pm$.7} \\
\midrule
GREASELM & Test & 39.3\tiny{$\pm$2.2} & 38.4\tiny{$\pm$.6} & 37.7\tiny{$\pm$1.4} & 38.5\tiny{$\pm$.8} & 38.1\tiny{$\pm$1.0} & 38.8\tiny{$\pm$.5} & 36.7\tiny{$\pm$.2} \\
\bottomrule[\lightrulewidth]
\end{tabular}
}
\caption{\textbf{End-to-end accuracy of two systems across UMLS constructions.} MedQA accuracy (\%), mean $\pm$ standard deviation over three seeds. Releases abbreviated (e.g., 20AA for 2020AA). \textit{Empty} keeps only the placeholder concepts used on linking failure.}
\label{tab:endtoend}
\end{table}  %
\section{Toward Better Use of UMLS}
\label{sec:tbu}

Taken together, these findings do not show that UMLS has been exhausted as a resource for MQA but rather that it remains underexplored.
A reported gain depends on the release and filters that define the available knowledge, the spans that survive \EL{}, the answer connections retained by \SR{}, and the form in which \KI{} exposes the result to the model.
The central question is therefore not whether UMLS is useful, but whether a documented construction makes UMLS evidence available to the decision, connects it to the candidate answer, and renders it intelligible to the model asked to use it.
\Cref{tab:reporting_checklist} condenses the preceding sections into the minimum information a UMLS-augmented MQA paper should report for such a construction to be documented.
We next discuss methodological insights for the effective use of UMLS and conduct an empirical evaluation to systematically quantify which controlled evidence representations are best suited for contemporary, reasoning-oriented LLM-based MQA frameworks.
Appendix~\ref{app:downstream_umls_details} reports an in-depth experimental setup.

\paragraph{Which UMLS evidence can reach the answer?}

UMLS contribution should be evaluated by whether retrieved evidence supports answer selection, rather than by question stem coverage alone.
On MedQA, question-side coverage would make MetaMap and scispaCy appear equivalent, since both ground 100.0\% of test stems.
MetaMap links at least one concept from the correct answer for 98.2\% of items and from the full answer set for 96.4\%, whereas scispaCy reaches 86.1\% and 65.0\%, respectively.
The relevant linker failure is therefore not absence from the clinical stem, but asymmetric loss of correct answers and distractors.
Downstream accuracy can be interpreted as useful UMLS injection only when question grounding, correct-answer grounding, and full-option grounding are reported jointly.

\paragraph{Which UMLS metadata should the reader see?}

A UMLS path is not medical evidence merely because it connects question and answer concepts: the reader must also be able to interpret what those concepts denote and why the relation is clinically meaningful.
Names and relation labels provide a route through the Metathesaurus but leave the biomedical role of the concepts underspecified, whereas definitions make each concept explicit, and semantic types indicate whether the path passes through a disease, finding, substance, or procedure.
Yet no single view recovers uniformly more baseline errors than the others (\cref{fig:evidence_form_model_sweep}).
Metadata-enriched paths are strongest on most MedQA and MedMCQA readers, whereas bare relation neighborhoods lead on nine of ten readers on PubMedQA; the ordering also moves across model families.
It moves again with the prompting regime: in a chain-of-thought pilot on MedQA, the winning evidence view changes relative to answer-only prompting (\cref{tab:prompting_regime}).
The evidence view at the \KI{} stage should therefore be treated as part of the intervention itself, specifying whether the reader receives only names and relations or also the metadata needed to turn a graph connection into usable biomedical evidence.

\begin{figure}[t]
    \centering
    \begin{subfigure}[t]{\linewidth}
        \centering
        \begin{tikzpicture}
            \scriptsize
            \node[draw=none, inner sep=1.5pt, align=center, text width=.96\linewidth] {
            \begin{tabular}{@{}lll@{}}
                \ref{plot:evidence_relation_only} \texttt{MRCONSO+MRREL} &
                \ref{plot:evidence_q_to_a_path} \texttt{MRCONSO+MRREL} $Q{\to}A$ &
                \ref{plot:evidence_metadata_rich} \texttt{+MRDEF+MRSTY}
            \end{tabular}
            };
        \end{tikzpicture}
    \end{subfigure}\\[1mm]
    \begin{subfigure}[t]{\linewidth}
        \centering
\begin{tikzpicture}
\begin{axis}[
    ybar,
    width=\linewidth,
    height=2.6cm,
    bar width=3.8pt,
    xmin=0.4, xmax=10.6,
    ymin=0, ymax=48,
    ytick={0,20,40},
    ylabel={MedQA},
    xtick=\empty,
    ymajorgrids=true,
    grid style={dashed, gray!45},
    axis background/.style={fill=plotbackground},
    axis on top,
    major tick length=0pt,
    ylabel style={font=\fontsize{7}{8}\selectfont},
    tick label style={font=\fontsize{6}{7}\selectfont},
    error bars/error bar style={line width=0.55pt, color=black},
    error bars/error mark options={
        rotate=90, mark size=2.0pt, line width=0.55pt, color=black
    },
]

\addplot+[
    fill=baselinecolor, draw=black, bar shift=-3.8pt,
    error bars/.cd, y dir=both, y explicit
] coordinates {
    (1, 22.0) +- (0, 4.0)
    (2, 24.0) +- (0, 4.7)
    (3, 19.4) +- (0, 3.9)
    (4, 13.2) +- (0, 3.0)
    (5, 16.7) +- (0, 4.2)
    (6, 23.4) +- (0, 3.4)
    (7, 29.7) +- (0, 4.1)
    (8, 18.8) +- (0, 5.4)
    (9, 18.8) +- (0, 5.4)
    (10, 20.0) +- (0, 5.7)
};
\label{plot:evidence_relation_only}

\addplot+[
    fill=darkerbaselinecolor, draw=black, bar shift=0pt,
    error bars/.cd, y dir=both, y explicit
] coordinates {
    (1, 26.0) +- (0, 4.3)
    (2, 27.0) +- (0, 4.9)
    (3, 22.6) +- (0, 4.1)
    (4, 18.4) +- (0, 3.5)
    (5, 20.8) +- (0, 4.5)
    (6, 21.3) +- (0, 3.3)
    (7, 27.0) +- (0, 4.0)
    (8, 12.5) +- (0, 4.5)
    (9, 31.2) +- (0, 6.4)
    (10, 26.7) +- (0, 6.3)
};
\label{plot:evidence_q_to_a_path}

\addplot+[
    fill=portscolor!80, draw=black, bar shift=3.8pt,
    error bars/.cd, y dir=both, y explicit
] coordinates {
    (1, 28.0) +- (0, 4.4)
    (2, 31.3) +- (0, 5.1)
    (3, 22.6) +- (0, 4.1)
    (4, 26.3) +- (0, 3.9)
    (5, 25.0) +- (0, 4.9)
    (6, 14.9) +- (0, 2.9)
    (7, 27.0) +- (0, 4.0)
    (8, 25.0) +- (0, 5.9)
    (9, 37.5) +- (0, 6.6)
    (10, 13.3) +- (0, 4.8)
};
\label{plot:evidence_metadata_rich}

\end{axis}
\end{tikzpicture}
\par\vspace{-1.2mm}\noindent
\begin{tikzpicture}
\begin{axis}[
    ybar,
    width=\linewidth,
    height=2.6cm,
    bar width=3.8pt,
    xmin=0.4, xmax=10.6,
    ymin=0, ymax=48,
    ytick={0,20,40},
    ylabel={MedMCQA},
    xtick=\empty,
    ymajorgrids=true,
    grid style={dashed, gray!45},
    axis background/.style={fill=plotbackground},
    axis on top,
    major tick length=0pt,
    ylabel style={font=\fontsize{7}{8}\selectfont},
    tick label style={font=\fontsize{6}{7}\selectfont},
    error bars/error bar style={line width=0.55pt, color=black},
    error bars/error mark options={
        rotate=90, mark size=2.0pt, line width=0.55pt, color=black
    },
]

\addplot+[
    fill=baselinecolor, draw=black, bar shift=-3.8pt,
    error bars/.cd, y dir=both, y explicit
] coordinates {
    (1, 20.5) +- (0, 2.0)
    (2, 28.6) +- (0, 3.0)
    (3, 21.4) +- (0, 2.3)
    (4, 21.4) +- (0, 1.9)
    (5, 21.7) +- (0, 2.6)
    (6, 28.1) +- (0, 1.8)
    (7, 24.3) +- (0, 2.1)
    (8, 11.1) +- (0, 2.2)
    (9, 9.1) +- (0, 2.6)
    (10, 22.9) +- (0, 2.2)
};

\addplot+[
    fill=darkerbaselinecolor, draw=black, bar shift=0pt,
    error bars/.cd, y dir=both, y explicit
] coordinates {
    (1, 25.6) +- (0, 2.1)
    (2, 14.3) +- (0, 2.3)
    (3, 21.4) +- (0, 2.3)
    (4, 14.3) +- (0, 1.6)
    (5, 26.1) +- (0, 2.8)
    (6, 21.1) +- (0, 1.6)
    (7, 13.5) +- (0, 1.7)
    (8, 27.8) +- (0, 3.2)
    (9, 9.1) +- (0, 2.6)
    (10, 17.1) +- (0, 1.9)
};

\addplot+[
    fill=portscolor!80, draw=black, bar shift=3.8pt,
    error bars/.cd, y dir=both, y explicit
] coordinates {
    (1, 25.6) +- (0, 2.1)
    (2, 14.3) +- (0, 2.3)
    (3, 25.0) +- (0, 2.5)
    (4, 26.2) +- (0, 2.1)
    (5, 30.4) +- (0, 2.9)
    (6, 33.3) +- (0, 1.9)
    (7, 18.9) +- (0, 2.0)
    (8, 22.2) +- (0, 3.0)
    (9, 18.2) +- (0, 3.5)
    (10, 25.7) +- (0, 2.2)
};

\end{axis}
\end{tikzpicture}
\par\vspace{-1.2mm}\noindent
\begin{tikzpicture}
\begin{axis}[
    ybar,
    width=\linewidth,
    height=2.6cm,
    bar width=3.8pt,
    xmin=0.4, xmax=10.6,
    ymin=0, ymax=48,
    ytick={0,20,40},
    ylabel={PubMedQA},
    xtick={1,2,3,4,5,6,7,8,9,10},
    xticklabels={
        {\qwenfigicon\,Qwen3.5-4B},
        {\qwenfigicon\,Qwen3.5-9B},
        {\llamafigicon\,Llama-3.1-8B},
        {\gemmafigicon\,MedGemma-4B},
        {\gemmafigicon\,MedGemma-27B},
        {\gemmafigicon\,Gemma-4-E2B},
        {\gemmafigicon\,Gemma-4-E4B},
        {\gemmafigicon\,Gemma-4-26B},
        {\gemmafigicon\,Gemma-4-31B},
        {\openaifigicon\,GPT-OSS-20B}
    },
    ymajorgrids=true,
    grid style={dashed, gray!45},
    axis background/.style={fill=plotbackground},
    axis on top,
    major tick length=0pt,
    ylabel style={font=\fontsize{7}{8}\selectfont},
    tick label style={font=\fontsize{6}{7}\selectfont},
    xtick style={draw=none},
    xticklabel style={rotate=40, anchor=east},
    error bars/error bar style={line width=0.55pt, color=black},
    error bars/error mark options={
        rotate=90, mark size=2.0pt, line width=0.55pt, color=black
    },
]

\addplot+[
    fill=baselinecolor, draw=black, bar shift=-1.9pt,
    error bars/.cd, y dir=both, y explicit
] coordinates {
    (1, 34.1) +- (0, 6.5)
    (2, 35.1) +- (0, 6.9)
    (3, 31.0) +- (0, 6.3)
    (4, 16.7) +- (0, 5.0)
    (5, 18.8) +- (0, 4.9)
    (6, 20.0) +- (0, 5.0)
    (7, 17.0) +- (0, 4.8)
    (8, 6.8) +- (0, 3.3)
    (9, 7.5) +- (0, 3.7)
    (10, 32.9) +- (0, 4.9)
};

\addplot+[
    fill=portscolor!80, draw=black, bar shift=1.9pt,
    error bars/.cd, y dir=both, y explicit
] coordinates {
    (1, 32.5) +- (0, 6.4)
    (2, 13.5) +- (0, 4.9)
    (3, 23.8) +- (0, 5.8)
    (4, 9.5) +- (0, 4.0)
    (5, 4.2) +- (0, 2.5)
    (6, 16.0) +- (0, 4.5)
    (7, 12.8) +- (0, 4.3)
    (8, 13.6) +- (0, 4.5)
    (9, 2.5) +- (0, 2.2)
    (10, 20.0) +- (0, 4.2)
};

\end{axis}
\end{tikzpicture} \space    \end{subfigure}
    \caption{\textbf{Record-level UMLS evidence contribution across readers and benchmarks.}
    Accuracy gain (+\%) and 95\% CI over zero-shot on the baseline-wrong subset, for 10 readers
    from 5 families under 3 \KI{}.}
    \label{fig:evidence_form_model_sweep}
\end{figure}
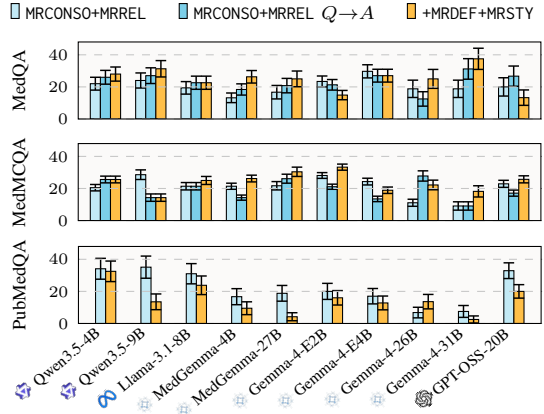 
\begin{table}[!h]
  \centering
  \small
  \setlength{\tabcolsep}{5pt}
  \renewcommand{\arraystretch}{1.15}
  \newcommand{\qwenicon}{\raisebox{-0.35em}{\includegraphics[height=1.38em]{qwen-icon.png}}\xspace}
  \newcommand{\llamaicon}{\raisebox{-0.09em}{\includegraphics[height=0.94em]{llama-icon.png}}\xspace}
  \newcommand{\gemmaicon}{\raisebox{-0.18em}{\includegraphics[height=1.12em]{gemma-icon.png}}\xspace}
  \newcommand{\openaiicon}{\raisebox{-0.10em}{\includegraphics[height=1.06em]{openai-icon.png}}\xspace}
    \begin{tabular*}{\columnwidth}{@{\extracolsep{\fill}}lrrr@{}}
      \toprule
      \multirow{2}{*}{\textbf{Model}}
        & \multicolumn{3}{c}{\textbf{Evidence view (+\%)}} \\
      \cmidrule(lr){2-4}
      & \textbf{Rel.} & \boldmath{$\mathcal{QA}$} & \textbf{+Meta} \\
      \midrule
      \qwenicon Qwen3.5-4B   & \best{25.7} & 22.9 & 22.9 \\
      \qwenicon Qwen3.5-9B   & 18.5 & 33.3 & \best{37.0} \\
      \llamaicon Llama-3.1-8B & \best{21.2} & 18.2 & 15.2 \\
      \gemmaicon MedGemma-4B  & 11.6 & 25.6 & \best{27.9} \\
      \gemmaicon MedGemma-27B & 15.4 & \best{23.1} & 19.2 \\
      \gemmaicon Gemma-4-E2B  & \best{25.5} & 23.5 & 17.6 \\
      \gemmaicon Gemma-4-E4B  & 29.3 & 29.3 & \best{31.7} \\
      \gemmaicon Gemma-4-26B  & 22.2 & 16.7 & \best{27.8} \\
      \gemmaicon Gemma-4-31B  & 17.6 & 29.4 & \best{47.1} \\
      \openaiicon GPT-OSS-20B  & 9.1 & \best{54.5} & 36.4 \\
      \bottomrule
    \end{tabular*}
  \caption{\textbf{Evidence view under chain-of-thought prompting.}
  Recovery rate (\%) on a chain-of-thought pilot over a subsample of the MedQA test split, under
  scispaCy linking, best view per row in \textbf{bold}. 
  \textit{Rel.}: \texttt{MRCONSO+MRREL} neighborhoods; \textit{$\mathcal{QA}$}: the same restricted to
  question-to-answer paths; \textit{+Meta}: paths with \texttt{MRDEF} definitions and
  \texttt{MRSTY} semantic types.}
  \label{tab:prompting_regime}
\end{table} 

\begin{figure}[t]
    \centering
    \begin{subfigure}[t]{\linewidth}
        \centering
        \begin{tikzpicture}
            \scriptsize
            \node[draw=none, inner sep=1.5pt, align=center, text width=.96\linewidth] {
            \begin{tabular}{@{}lll@{}}
                \ref{plot:path_depth_two} 2 hops &
                \ref{plot:path_depth_three} 3 hops &
                \ref{plot:path_depth_four} 4 hops
            \end{tabular}
            };
        \end{tikzpicture}
    \end{subfigure}\\[-1mm]
    \begin{subfigure}[t]{\linewidth}
        \centering
        \begin{tikzpicture}
        \begin{axis}[
            ybar,
            clip=false,
            width=\linewidth,
            height=4.3cm,
            bar width=3.7pt,
            ymin=0,
            ymax=80,
            ylabel={Accuracy gain (+\%)},
            xlabel={Qwen3.5 model scale},
            xtick={1,2,3,4,5,6},
            xticklabels={0.8B,2B,4B,9B,27B,35B},
            ytick={0,20,40,60},
            ymajorgrids=true,
            grid style={dashed, gray!45},
            axis background/.style={fill=plotbackground},
            enlarge x limits=0.10,
            major tick length=0pt,
            title style={font=\fontsize{8}{8}\selectfont},
            ylabel style={font=\fontsize{7}{8}\selectfont},
            xlabel style={font=\fontsize{7}{8}\selectfont},
            tick label style={font=\fontsize{6}{7}\selectfont},
            axis on top,
            error bars/error bar style={line width=0.55pt, color=black},
            error bars/error mark options={
                rotate=90, mark size=2.0pt, line width=0.55pt, color=black
            },
        ]

        \addplot+[
            fill=condPath, draw=black,
            error bars/.cd, y dir=both, y explicit
        ] coordinates {
            (1, 19.2) +- (0, 5.1)
            (2, 27.3) +- (0, 6.7)
            (3, 31.8) +- (0, 8.2)
            (4, 33.6) +- (0, 9.4)
            (5, 48.0) +- (0, 11.8)
            (6, 52.2) +- (0, 12.9)
        };
        \label{plot:path_depth_two}

        \addplot+[
            fill=portscolor!80, draw=black,
            error bars/.cd, y dir=both, y explicit
        ] coordinates {
            (1, 19.9) +- (0, 5.2)
            (2, 29.7) +- (0, 6.9)
            (3, 33.3) +- (0, 8.3)
            (4, 38.2) +- (0, 9.7)
            (5, 51.9) +- (0, 11.8)
            (6, 53.5) +- (0, 12.8)
        };
        \label{plot:path_depth_three}

        \addplot+[
            fill=taclRed, draw=black,
            error bars/.cd, y dir=both, y explicit
        ] coordinates {
            (1, 31.8) +- (0, 6.1)
            (2, 33.1) +- (0, 7.1)
            (3, 35.8) +- (0, 8.5)
            (4, 29.6) +- (0, 9.1)
            (5, 50.0) +- (0, 11.8)
            (6, 54.5) +- (0, 12.8)
        };
        \label{plot:path_depth_four}

        \end{axis}
        \end{tikzpicture}
    \end{subfigure}
    \caption{\textbf{Question-to-answer path length effect on the \qwenfigicon~Qwen3.5 ladder.} Accuracy gain (+\%) and 95\% CI on MedQA over zero-shot across \SR{} configurations. Each $k$ setting uses
    only pairs where every answer option is reachable from a question node by an exact $k$-hop path.}
    \label{fig:path_depth_model_sweep}
\end{figure}
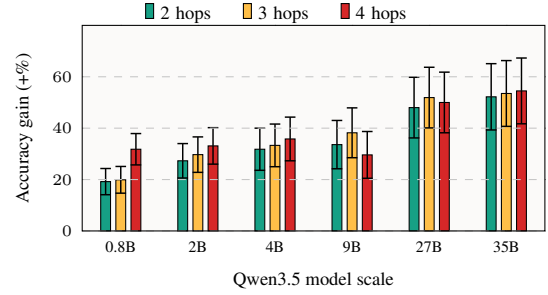

\paragraph{How should the reader navigate UMLS paths?}
The shortest UMLS path should not be assumed to be the most useful evidence path.
In GNN settings, shorter graph distances are often preferred to mitigate oversmoothing, but question-to-answer path depth plays a different role when the path is exposed as structured evidence.
Within the Qwen3.5 scale ladder on MedQA, deeper question-to-answer paths yield stronger gains under matched evidence conditions, suggesting that additional intermediate concepts help the reader distinguish an answer supported by a biomedical chain from distractors that are merely close in the ontology (\cref{fig:path_depth_model_sweep}).
The design goal in \SR{} should thus not be to minimize graph distance in isolation from the outset, but to surface paths that stay centered on the question-to-answer decision while offering sufficient structure for the reader to treat them as evidence.

\section{Conclusion}
\label{sec:conclusion}

This survey has argued that current work on UMLS-based medical question answering is harder to interpret than benchmark tables suggest. 
Across the literature, UMLS is rarely used as a stable resource; 
instead, it is repeatedly reshaped through choices in source selection, entity linking, graph construction, and downstream integration. 
Reported gains are therefore often compared across whole pipelines rather than isolating the contribution of UMLS itself.
The answer suggested by this review is thus a qualified maybe not: we are not yet done with UMLS injection for medical question answering, not because its value has already been clearly established, but because the field has not yet tested it under conditions that make its contribution identifiable. 
What remains is not simply to build new methods, but to evaluate UMLS more directly and more reproducibly, so that both its limits and its potential can be assessed with greater clarity. %
\section*{Limitations}

This paper asks whether the published record permits a reported gain to be attributed to UMLS itself rather than to the pipeline that reshapes it.
We answer with a systematic review of the studies meeting our inclusion criteria, namely peer-reviewed work from the last five years in which UMLS plays a substantive role in a medical question answering or closely related reasoning setting, and whose methodology is documented in enough detail to be compared across systems.
Database, subject-area, language, access, and date restrictions can exclude relevant work, including clinical-informatics studies.
Our review is also bounded by what authors chose to report: where a paper is silent, we recover the missing fields from released code and supplementary material, and mark them explicitly when neither source resolves them (\cref{app:review_protocol}).

We complement the review with controlled experiments that isolate one construction factor at a time.
What none of them establishes is external validity: we do not claim that the measured orderings transfer to medical QA datasets, entity linkers, or model families outside those we evaluate.
Our reader experiments show a recovery rate derived from items answered incorrectly, but this does not support conclusions about which evidence view most effectively enhances overall accuracy.
Our linking measurements are primarily counts of reachability. 
A manual audit of answer options under UMLS 2023AA confirms that MetaMap's higher yield carries substantial noise, but leaves linking precision on question stems and other benchmarks unmeasured; it also judges whether a concept matches its span, not whether that concept helps a downstream model. 
Our controlled comparisons draw on \texttt{MRCONSO}, \texttt{MRREL}, \texttt{MRDEF}, and \texttt{MRSTY}, leaving the remaining Metathesaurus tables and the SPECIALIST Lexicon unexamined.

Our claim concerns whether a reported gain can be traced to UMLS, not how large the effect is in any individual published system.
MQA benchmarks do not capture the temporality, uncertainty, and accountability of real clinical decision-making, and our results should not be read as validating UMLS-augmented LMs for clinical deployment without further evaluation. %
\section*{Acknowledgements}

Research partially supported by: \href{https://aipact-edih.it}{AI-PACT} (CUP B47H22004450008, B47H22004460001); National Plan PNC-I.1 \href{https://www.fondazionedare.it/en/progetto-obiettivi-struttura/}{DARE} (PNC0000002, CUP B53C22006450001); 2024 Scientific Research and High Technology Program, project ``\href{https://disi-unibo-nlp.github.io/projects/carisbo/}{AI analysis for risk assessment of empty lymph nodes in endometrial cancer surgery}'', the Fondazione Cassa di Risparmio in Bologna; Horizon Europe VALWARP project (proposal no. 101326664). 

\bibliography{references}

\clearpage
\newpage

\appendix
\section{UMLS License and Data Usage}
\label{app:umls_license}

The UMLS Metathesaurus is distributed by the NLM under a free, nonexclusive, non-transferable license granted to registered users upon completion of a brief online agreement.\footnote{\url{https://uts.nlm.nih.gov/uts/signup-login}}
The license permits use in research applications and computer systems designed to improve access to biomedical information, subject to three principal restrictions: (i) the Metathesaurus or any subset of it---including individual source vocabularies---may not be redistributed as a standalone artifact; (ii) LICENSEE must acknowledge NLM as the source, citing the release version, without implying NLM endorsement; and (iii) some integrated source vocabularies carry additional per-vocabulary restrictions, ranging from prohibition on derivative works to restriction of use to internal research only.
We obtained access under this license for research purposes.
We use the 2025AB release for the resource-level analysis of \cref{sec:what_umls_is,sec:what_umls_has_become}, for the expansion experiments of \cref{sec:subgraph}, and for the evidence constructions of \cref{sec:tbu}; the release-sensitivity experiments of \cref{sec:nel,sec:injection} additionally use the 2020AA, 2021AA, 2022AB, and 2023AA releases, each obtained under the same license.
No UMLS data, subsets, or derived graph artifacts are released 
or redistributed as part of this paper.

\section{Screening Protocol and Paper-Level Annotations}
\label{app:review_protocol}

\subsection{Search Strategy on Scopus}
We conducted a structured search on Scopus on January~15,~2026.
We restricted results to English-language journal articles and conference papers.
A representative search string is given in \cref{fig:literature_search_query}.

\begin{figure}[h!]
\centering
\small
\begin{tcolorbox}[
    colframe=black,
    colback=gray!5,
    coltitle=white,
    colbacktitle=black,
    title=Literature Search Query,
    fonttitle=\bfseries,
    boxrule=0.8pt,
    arc=2mm,
    left=2mm,
    right=2mm,
    top=2mm,
    bottom=2mm
]
\begin{verbatim}
ABS (
    ("*knowledge graph*")
    OR ("knowledge base*")
    OR ("structured")
    OR ("graph*")
)
AND ABS (
    ("natural language processing")
    OR ("NLP")
    OR ("language model*")
    OR ("LLM*")
    OR ("LM*")
)
AND ABS (
    ("question answering")
    OR ("*QA*")
    OR ("reasoning*")
)
AND PUBYEAR > 2019
AND PUBYEAR < 2026

AND LIMIT-TO ( SRCTYPE,  "p" )
AND LIMIT-TO ( PUBSTAGE, "final" )

AND (
    LIMIT-TO ( DOCTYPE, "cp" )
    OR LIMIT-TO ( DOCTYPE, "ar" )
)

AND LIMIT-TO ( SUBJAREA, "COMP" )
AND LIMIT-TO ( LANGUAGE, "English" )
\end{verbatim}
\end{tcolorbox}
\caption{\textbf{Structured search query.} Used to retrieve literature on knowledge graphs and language models for question answering and reasoning from Scopus.}
\label{fig:literature_search_query}
\end{figure}

\subsection{Screening and Eligibility}
Retrieved records were screened in two stages: title/abstract screening followed by full-text assessment. 
We retained studies only when UMLS (or a clearly UMLS-derived resource) played a substantive technical role in the method, rather than being mentioned only as background or used merely as one benchmark knowledge graph among many. 
We further required the paper to address question answering or closely related reasoning-oriented biomedical NLP settings, and to provide enough methodological detail for comparison across knowledge grounding, retrieval, and injection stages.
At full-text screening, we excluded papers whose primary focus fell outside the scope of UMLS-grounded reasoning systems for MQA.

\subsection{Utilized QA Benchmark Datasets}
\label{app:benchmark_datasets}
The reviewed studies adopt heterogeneous evaluation setups, including biomedical textual entailment, clinical NER, and other domain-specific tasks. 
Here we focus on the subset that evaluates on standard QA benchmarks, which recur most frequently across the corpus and form the basis of the cross-system comparisons discussed in \S\ref{sec:how_umls_is_used}. 
These benchmarks fall into two evaluation paradigms: \textit{closed-form benchmarks}, which require selection from a fixed answer set (yes/no/maybe or multiple choice), and \textit{open-form benchmarks}, which require generating a free-text response. 
\paragraph{Closed-form Benchmarks}
\begin{itemize}
    \item \textbf{PubMedQA}~\citep{DBLP:conf/emnlp/JinDLCL19}: a biomedical QA dataset in which models answer research questions with ``yes,'' ``no,'' or ``maybe,'' grounded in the corresponding PubMed abstract.
    \item \textbf{BioASQ}~\citep{DBLP:journals/corr/abs-2006-09174}: a benchmark suite for large-scale biomedical semantic indexing and question answering, covering yes/no, factoid, list, and summary question types over biomedical literature.
    \item \textbf{MedQA (USMLE)}~\citep{DBLP:journals/corr/abs-2009-13081}: an open-domain multiple-choice dataset compiled from professional medical board exams in English and Chinese, designed to test clinical reasoning.
    \item \textbf{MedMCQA}~\citep{DBLP:conf/chil/PalUS22}: over 194{,}000 real-world multiple-choice questions from Indian medical entrance exams, covering a wide range of medical subjects.
\end{itemize}

\paragraph{Open-form Benchmarks}
\begin{itemize}
    \item \textbf{MedicationQA}~\citep{DBLP:conf/medinfo/AbachaMSGSD19}: a gold-standard corpus of real consumer health questions about medications, annotated with question type, focus, and trusted answer sources.
    \item \textbf{COVID-QA}~\citep{DBLP:conf/emnlp/2020covid}: over 2{,}000 question--answer pairs annotated by biomedical experts on COVID-19-related scientific articles.
    \item \textbf{ExpertQA}~\citep{DBLP:conf/naacl/MalaviyaLCSYR24}: long-form, information-seeking questions curated by domain experts across 32 fields, paired with expert-verified model-generated answers and per-claim attributions.
\end{itemize}
Representative items from each dataset are shown in Table~\ref{tab:qa_examples}.
\definecolor{answerblue}{RGB}{0, 71, 171}
\definecolor{answergreen}{HTML}{1B5E20} 

\begin{table*}[htbp]
    \centering
    \small
    \setlength{\tabcolsep}{4pt} 
    \renewcommand{\arraystretch}{1.1}
    \begin{tabular}{l p{12.5cm}}
        \toprule
        \textbf{Dataset} & \textbf{Example} \\
        \midrule
        
        BioASQ & Is the monoclonal antibody Trastuzumab (Herceptin) of potential use in the treatment of prostate cancer? \newline 
        \textcolor{answergreen}{\textbf{(A) Yes}} \quad (B) No \\
        \midrule
        
        PubMedQA & Are group 2 innate lymphoid cells ( ILC2s ) increased in chronic rhinosinusitis with nasal polyps or eosinophilia? \newline 
        \textcolor{answergreen}{\textbf{(A) Yes}} \quad (B) No \\
        \midrule
        
        MedQA & The only immunoglobulin found as a dimer has what primary function? \newline 
        (A) Protect against invasive helminth infection? \newline
        (B) Protect against viral infections \newline 
        \textcolor{answergreen}{\textbf{(C) Inhibiting bacterial adherance and colonization of mucous membranes}} \newline 
        (D) Provides the most specific recognition to circulating antigens in the bloodstream \\
        \midrule
        
        MedMCQA & Chronic urethral obstruction due to benign prismatic hyperplasia can lead to the following change in kidney parenchyma? \newline 
        (A) Hyperplasia \quad (B) Hyperophy \quad \textcolor{answergreen}{\textbf{(C) Atrophy}} \quad (D) Dyplasia \\
        \midrule
        
        MedicationQA & what meloxicam look like? \newline 
        \textcolor{answergreen}{\textbf{Product Characteristics Color YELLOW (light yellow) Score no score Shape OVAL Size 3mm Imprint Code S160}} \\
        \midrule
        
        CovidQA & What plays the crucial role in the Mother to Child Transmission of HIV-1 and what increases the risk? \newline 
        \textcolor{answergreen}{\textbf{DC-SIGNR plays a crucial role in MTCT of HIV-1 and that impaired placental DC-SIGNR expression increases risk of transmission.}} \\
        \midrule
        
        ExpertQA & What does EPRR stand for? \newline 
        \textcolor{answergreen}{\textbf{EPRR stands for Emergency Preparedness, Resilience and Response}} \\
        
        \bottomrule
    \end{tabular}
    \caption{\textbf{Verbatim representative items from each QA benchmark used in the reviewed studies.} Correct or reference answers are highlighted in \textcolor{answergreen}{\textbf{green}}.}
    \label{tab:qa_examples}
\end{table*}

\subsection{Per-paper Annotations}
\label{app:annotations_per_paper}
Table~\ref{tab:annotations_per_paper} summarizes the operationalization of UMLS in each of the 25 reviewed systems along the dimensions discussed in \S\ref{sec:how_umls_is_used}.
Among them, six studies meet our operational reproducibility criterion: released code and recoverable graph scale, entity linker, expansion depth $K$, and injection mechanism. 
We resolved fields from the paper and supplement, then released code, then code from cited systems.
For each system, we report: the publication venue; the UMLS release used (\textbf{UMLS Ver.}, NA if unreported);
the consulted UMLS tables; any subset or filter applied (source vocabulary, relation, semantic type, or language); the resulting graph scale (\textbf{UMLS Size}); the named entity linker (\textbf{NEL});
the subgraph expansion depth and node-budget cap (\textbf{$K$\,/\,Budget}); and the injection paradigm from \S\ref{sec:injection} --- textual linearization; structural injection; parameter encoding.
Entries marked NA indicate that the corresponding detail is not resolvable from the paper, supplement, or released code.
A GitHub icon (\faGithub) next to a system name indicates that the authors have released code;
clicking the icon opens the repository.
The full per-paper annotation, including evaluation datasets and qualitative notes, is available in our supplementary repository.\footnote{\url{https://github.com/disi-unibo-nlp/are-we-done-with-umls}}

\renewcommand{\arraystretch}{1.1}
\setlength{\tabcolsep}{4pt}

\begin{sidewaystable*}[!htbp]
\centering
\footnotesize
\caption{\textbf{Per-paper annotations of UMLS operationalization across the reviewed systems.}}
\label{tab:annotations_per_paper}

\resizebox{0.85\textheight}{!}{%
\begin{tabular}{@{}r 
>{\raggedright\arraybackslash}p{5.2cm} l l 
>{\raggedright\arraybackslash}p{2.4cm} 
>{\raggedright\arraybackslash}p{2.4cm} 
>{\raggedright\arraybackslash}p{3.4cm} 
>{\raggedright\arraybackslash}p{1.8cm} 
>{\raggedright\arraybackslash}p{2.4cm} 
>{\raggedright\arraybackslash}p{1.7cm}@{}}

\toprule
\textbf{\#} 
& \textbf{System} 
& \textbf{Venue} 
& \textbf{UMLS Ver.} 
& \textbf{UMLS Tables} 
& \textbf{UMLS Filter} 
& \textbf{UMLS Size} 
& \textbf{NEL} 
& \textbf{$K$/ Budget} 
& \textbf{Injection mechanism choices} \\
\midrule

1  & DRAGON~\citep{DBLP:conf/nips/YasunagaBR0MLL22}\,\href{https://github.com/michiyasunaga/dragon}{\faGithub}
   & NeurIPS '22  
   & NA 
   & MRCONSO, MRREL  
   & NA 
   & 300K nodes; 1M edges                                                 
   & scispaCy   
   & $K{=}2$ / 200 nodes
   & Structural injection \\

2  & MediSage~\citep{DBLP:conf/www/ChoudhuryAHT23}
   & WWW '23      
   & NA       
   & MRCONSO, MRREL, MRSTY  
   & NA 
   & NA
   & string match 
   & $K{=}3$ / NA
   & NA \\

3  & KnowledgeDA~\citep{DBLP:conf/acl/Ding0W23}\,\href{https://github.com/RuiqingDing/KnowledgeDA}{\faGithub}
   & Findings ACL '23 
   & 2021AA 
   & MRCONSO, MRREL, MRSTY  
   & non-Eng.\ 502 rel. types \ medicine related sem. types
   & 1.15M CUIs; 4.88M triples           
   & neural embedding 
   & $K{=}2$ / NA
   & Textual linearization \\

4  & KG-Infused LLM~\citep{DBLP:conf/icmla/KattaRFC24}\,\href{https://github.com/SivaKatta1010/NeoKG-LLM}{\faGithub}
   & ICMLA '24    
   & NA       
   & MRCONSO, MRREL, MRDEF 
   & MSH and NCI
   & 622K nodes / 4.94M edges 
   & scispaCy   
   & $K{=}1$ / 20 relations
   & Textual linearization \\

5  & QAVSA~\citep{laube-eliasmith-2024-qavsa}
   & RepL4NLP '24 
   & NA       
   & MRCONSO, MRREL  
   & NA 
   & 300K nodes; 1M edges    
   & scispaCy  
   & $K{=}2$ / NA       
   & Structural injection \\

6  & QAT~\citep{DBLP:conf/aaai/ParkCKPKJKK23}
   & AAAI '23     
   & NA       
   & MRCONSO, MRREL  
   & DDB + DrugBank
   & 9K nodes; 44K edges       
   & NA                                            
   & $K{=}2$ / 200 nodes  
   & Structural injection \\

7  & QA-GNN~\citep{DBLP:conf/naacl/YasunagaRBLL21}\,\href{https://github.com/michiyasunaga/qagnn}{\faGithub}
   & NAACL '21    
   & NA       
   & MRCONSO, MRREL 
   & DDB + DrugBank
   & 9K nodes; 44K edges
   & scispaCy   
   & $K{=}2$ / 200 nodes  
   & Structural injection \\

8  & \citet{DBLP:conf/bibm/SenguptaHCSM24}\,\href{https://github.com/saptarshi059/CDQA-project}{\faGithub}
   & BIBM '24     
   & NA       
   & MRCONSO, MRREL, MRDEF
   & NA 
   & NA                                                                                 
   & MetaMap    
   & $K{=}1$ / NA       
   & Structural injection \\

9  & GNP~\citep{DBLP:conf/aaai/0001SWWHWCX24}
   & AAAI '24     
   & NA       
   & MRCONSO, MRREL
   & NA 
   & NA                           
   & scispaCy   
   & $K{=}2$ / NA       
   & Structural injection \\

10 & \citet{DBLP:conf/icaart/VladikaFM24}
   & ICAART '24   
   & 2023AA   
   & MRCONSO, MRREL
   & SNOMED CT, U.S.
   & 1.75M triplets                                
   & NA         
   & NA / NA            
   & Parameter encoding \\

11 & \citet{DBLP:conf/coling/YadavPS20}\,\href{https://github.com/VishalPallagani/Medical-Knowledge-enriched-Textual-Entailment}{\faGithub}
   & COLING '20   
   & NA       
   & MRCONSO, MRREL
   & NA 
   & NA                                 
   & scispaCy   
   & $K{=}2$ / NA       
   & Structural injection \\

12 & KG-Rank~\citep{DBLP:conf/bionlp/YangLMZKLCCCML24}\,\href{https://github.com/YangRui525/KG-Rank}{\faGithub}
   & BioNLP '24   
   & NA       
   & MRCONSO, MRREL 
   & NA 
   & NA                                             
   & LLM prompt 
   & $K{=}1$ / NA       
   & Textual linearization \\

13 & DRLK~\citep{DBLP:conf/emnlp/ZhangD0H22}\,\href{https://github.com/MZ-MiaoZhang/DRLK}{\faGithub}
   & EMNLP '22    
   & NA       
   & MRCONSO, MRREL
   & NA 
   & 9K nodes; 44K edges                                                
   & scispaCy   
   & $K{=}2$ / 200 nodes
   & Structural injection \\

14 & GREASELM~\citep{DBLP:conf/iclr/0001BYRLML22}\,\href{https://github.com/snap-stanford/GreaseLM}{\faGithub}
   & ICLR '22     
   & NA       
   & MRCONSO, MRREL  
   & DDB + DrugBank
   & 9K nodes; 44K edges                   
   & scispaCy   
   & $K{=}2$ / 200 nodes 
   & Structural injection \\

15 & \citet{DBLP:conf/www/0006D0Y24}
   & WWW '24      
   & NA       
   & MRCONSO, MRREL 
   & DDB
   & 50K triples                                   
   & neural embedding 
   & $K{=}1$ / NA
   & Textual linearization \\

16 & InfuserKI~\citep{DBLP:conf/emnlp/WangBWYLCC24}
   & EMNLP '24    
   & NA       
   & MRCONSO, MRREL
   & NA 
   & 2.5K \& 25K triplets                          
   & NA         
   & NA / NA         
   & Parameter encoding \\

17 & HiPrompt~\citep{DBLP:conf/emnlp/LuoVPH23}
   & SIGIR '23    
   & NA       
   & MRCONSO
   & NA 
   & NA                                      
   & NA                                      
   & NA / NA                      
   & NA \\

18 & UMLS-KGI-BERT~\citep{DBLP:conf/acl-clinicalnlp/MannionSG23}
   & ClinicalNLP '23 
   & 2022AB 
   & MRCONSO, MRREL, MRSTY 
   & human bioloy and medicine related sem. types
   & 2.53M CUIs
   & string match 
   & $K{=}1$ / NA
   & Parameter encoding \\

19 & Healthcare QA~\citep{DBLP:conf/icis/SukhwalKR22}
   & ICIS '22     
   & NA       
   & MRCONSO
   & NA 
   & 7.17K CUIs                                  
   & scispaCy   
   & $K{=}1$ / NA      
   & NA \\

20 & Chest ImaGenome~\citep{DBLP:conf/nips/WuALLSPYDMKGCM21}
   & NeurIPS '21  
   & NA       
   & MRCONSO
   & NA 
   & NA                                                                                
   & scispaCy   
   & NA / NA            
   & NA \\

21 & GLaM~\citep{DBLP:conf/aaaiss/DernbachAZHC24}
   & AAAI-SSS '24 
   & NA       
   & MRCONSO, MRREL 
   & 4 rel. types
   & 126K triples 
   & NA 
   & NA 
   & Textual linearization \\

22 & \citet{DBLP:conf/emnlp/ParkJKK24}
   & other        
   & NA       
   & MRCONSO, MRREL, MRSTY, SRDEF\textsuperscript{\dag}
   & NA 
   & NA                            
   & NA                                                 
   & NA / NA          
   & Parameter encoding \\

23 & MoP~\citep{DBLP:conf/emnlp/Meng0CSC21}\,\href{https://github.com/cambridgeltl/mop}{\faGithub}
   & EMNLP '21    
   & NA       
   & MRCONSO, MRREL 
   & SNOMED CT, US
   & NA                                            
   & NA         
   & NA / NA           
   & Parameter encoding \\

24 & UmlsBERT~\citep{DBLP:conf/naacl/MichalopoulosWK21}\,\href{https://github.com/gmichalo/UmlsBER}{\faGithub}
   & NAACL '21    
   & 2020AA   
   & MRCONSO, MRSTY
   & NA 
   & NA                                                                                
   & cTAKES     
   & NA / NA           
   & Parameter encoding \\

25 & DAKI~\citep{DBLP:conf/emnlp/LuDN21}
   & Findings EMNLP '21 
   & NA 
   & MRCONSO, MRREL
   & NA 
   & 1.77M triples                                 
   & NA         
   & $K{=}1$ / NA       
   & Parameter encoding \\
\bottomrule

\multicolumn{9}{@{}l}{\strut\small\textsuperscript{\dag}SRDEF is part of the UMLS Semantic Network rather than the Metathesaurus.} \\

\end{tabular}%
} %
\end{sidewaystable*}

\section{Minimum Reporting Checklist}
\label{app:checklist}

\Cref{tab:reporting_checklist} lists the construction choices that, as shown in \cref{sec:what_umls_has_become,sec:how_umls_is_used}, change the UMLS evidence reaching the model and should therefore be reported.

\begin{table}[t]
\centering
\fontsize{8.5}{10}\selectfont
\begin{tabularx}{\columnwidth}{@{}l>{\raggedright\arraybackslash}Xl@{}}
\toprule
\textbf{Stage} & \textbf{Report} & \textbf{See} \\
\midrule
\multirow{2}{*}{UMLS}
  & $\square$ Release version & \S\ref{sec:what_umls_has_become} \\
  & $\square$ Subset: source vocabularies and filters on the concepts and relations & \S\ref{sec:what_umls_has_become}, \ref{sec:nel} \\
  & $\square$ Metadata: fields attached to each concept, e.g.\ definitions, semantic types & \S\ref{sec:what_umls_is}, \ref{sec:tbu} \\
\midrule
\multirow{2}{*}{\EL{}}
  & $\square$ Linker: tool, non-default settings & \S\ref{sec:nel} \\
  & $\square$ Linking result: entity coverage, unlinked items & App.~\ref{app:linker_stats}, \ref{app:appendix_linking_coverage} \\
\midrule
\multirow{5}{*}{\SR{}}
  & $\square$ Connectivity rule & \S\ref{sec:subgraph} \\
  & $\square$ Depth $K$ and subgraph scale & \S\ref{sec:subgraph} \\
  & $\square$ Pruning criterion and node budget & App.~\ref{app:pruning_details} \\
  & $\square$ Fallback when linking yields no seed & \S\ref{sec:subgraph} \\
\bottomrule
\end{tabularx}
\caption{\textbf{Minimum reporting checklist for UMLS-augmented MQA.} Each item changes the UMLS evidence reaching the model and should be reported for a gain to be attributable to UMLS.}
\label{tab:reporting_checklist}
\end{table} 
\section{Linker Statistics on MedQA Dev Split}
\label{app:linker_stats}

We computed per-question entity counts on the MedQA dev split ($n{=}1{,}272$) for scispaCy and MetaMap (\cref{tab:per_side_entity}). 
For answer options, we report the union of entities linked across the four options. 
\texttt{Diff(M-S)} denotes the per-question difference (MetaMap minus scispaCy).

\begin{table}[h]
\centering
\resizebox{\columnwidth}{!}{
\begin{tabular}{@{}llrrrrr@{}}
\toprule
\textbf{Span} & \textbf{System} & \textbf{Mean} & \textbf{SD} & \textbf{Min} & \textbf{Max} & \textbf{Var} \\
\midrule
\multirow{3}{*}{Question stems} 
  & MetaMap    & 113.57 & 42.68 &   3 & 334 & 1821.34 \\
  & scispaCy   &  57.67 & 22.67 &   1 & 172 &  514.00 \\
  & Diff (M-S) &  55.90 & 25.02 &   1 & 164 &  626.06 \\
\midrule
\multirow{3}{*}{Answer options} 
  & MetaMap    &  13.82 & 11.46 &   0 & 109 &  131.29 \\
  & scispaCy   &   9.96 &  6.33 &   0 &  45 &   40.02 \\
  & Diff (M-S) &   3.86 &  8.61 & $-$20 &  64 &   74.12 \\
\bottomrule
\end{tabular}
}
\caption{\textbf{Per-question entity counts on MedQA dev ($n{=}1{,}272$) using UMLS 2020AA.} Answer-option counts
are the union of the concepts linked across the four options, not their sum.
\texttt{Diff(M-S)} gives the per-question difference between MetaMap and scispaCy.}
\label{tab:per_side_entity}
\end{table}

The \texttt{Diff(M-S)} row shows that on question stems MetaMap returns more entities than scispaCy for every question ($\min{=}1$), while on answer options scispaCy occasionally returns more ($\min{=}-20$), indicating that the option-side coverage gap between linkers is less consistent than the stem-side gap.

\paragraph{Counting Protocol}

We count one CUI per linked span, taking MetaMap's top-ranked candidate mapping rather than all candidates, and deduplicate CUIs within an instance so that repeated mentions of the same concept count once. Overlapping spans are resolved by retaining the longest match. Answer-option counts are the union across the four options, not their sum. We apply no semantic-type restriction, no negation detection, and no abbreviation expansion beyond each tool's defaults, so the counts reflect each linker as surveyed systems use it out of the box.

\section{Detailed Linking Coverage Statistics}
\label{app:appendix_linking_coverage}

\Cref{tab:missing_entities} reports split-level linking coverage for the three UMLS-based MQA systems analyzed in \cref{sec:nel} and for our MetaMap processing of the same MedQA splits.
For each system and split, a field is marked \textit{Present} if at least one UMLS concept was linked and \textit{Empty} otherwise; the \textit{Both Empty} column counts instances where neither the question stem nor the answer-option set produced any linked concept.
QA-GNN and GREASELM---both using scispaCy---leave the answer side unlinked in 59\% of test instances, while DRAGON reduces this to 24\% through a broader source vocabulary.
MetaMap reduces answer-side failures to under 1\% (44 test instances), with no question stem left entirely unlinked across any split, confirming that the coverage gap is a consequence of linker choice rather than an inherent property of the benchmark.

\begin{table}[t]
  \centering
  \small
  \resizebox{\linewidth}{!}{
  \begin{tabular}{llrrrrc}
    \toprule
    \multirow{2}{*}{\textbf{System}} & \multirow{2}{*}{\textbf{Split}} & \multicolumn{2}{c}{\textbf{Answer-side Links}} & \multicolumn{2}{c}{\textbf{Question-side Links}} & \multirow{2}{*}{\textbf{Both Empty}} \\
    \cmidrule(lr){3-4} \cmidrule(lr){5-6}
    & & Present & \textbf{Empty} & Present & \textbf{Empty} & \\
    \midrule
    \multirow{3}{*}{QA-GNN}
    & Dev   & 2,056  & 3,032  & 4,872  & 216  & 148 \\
    & Test  & 2,096  & 2,996  & 4,924  & 168  & 119 \\
    & Train & 16,115 & 24,597 & 38,836 & 1,876 & 1,320 \\
    \midrule
    \multirow{3}{*}{GREASELM}
    & Dev   & 2,055  & 3,033  & 4,872  & 216  & 148 \\
    & Test  & 2,090  & 3,002  & 4,924  & 168  & 119 \\
    & Train & 16,175 & 24,537 & 38,888 & 1,824 & 1,286 \\
    \midrule
    \multirow{3}{*}{DRAGON}
    & Dev   & 3,874  & 1,214  & 5,084  & 4    & 0    \\
    & Test  & 3,845  & 1,247  & 5,088  & 4    & 4    \\
    & Train & 30,612 & 10,100 & 40,679 & 33   & 9    \\
    \midrule
    \multirow{3}{*}{MetaMap}
    & Dev   & 5,036  & 52  & 5,088  & 0    & 0    \\
    & Test  & 5,048  & 44  & 5,092  & 0    & 0    \\
    & Train & 40,280 & 432 & 40,712 & 0   & 0    \\
    \bottomrule
  \end{tabular}
  }
  \caption{\textbf{Linking coverage on MedQA preprocessing artifacts from QA-GNN, GREASELM, and DRAGON, and our MetaMap processing.} \textit{Present}: at least one UMLS concept linked. \textit{Empty}: no concept linked. \textit{Both Empty}: neither question stem nor answer-option set produced any linked concept.}
  \label{tab:missing_entities}
\end{table}

\section{Pruning and Evidence Loss}
\label{app:pruning_details}

We inspected the publicly released preprocessing and data-loading code of three UMLS-based MQA systems---QA-GNN~\citep{DBLP:conf/naacl/YasunagaRBLL21}, GREASELM~\citep{DBLP:conf/iclr/0001BYRLML22}, and DRAGON~\citep{DBLP:conf/nips/YasunagaBR0MLL22}---examining the 
latest \texttt{main} branches at the time of analysis.\footnote{QA-GNN: \url{https://github.com/michiyasunaga/qagnn}, GREASELM: \url{https://github.com/snap-stanford/GreaseLM}, DRAGON: \url{https://github.com/michiyasunaga/dragon}}
All three share the same structural pipeline: score bridge nodes by relevance to the QA context, retain the top-scoring candidates alongside the originally linked concepts, and truncate the result to a fixed node budget of 200 (\cref{tab:per_system_procedure}).

The dominant scoring strategy concatenates each candidate entity mention with the question-answer pair and computes a token likelihood under a pre-trained LM---specifically RoBERTa-large under a masked language modeling objective~\citep{DBLP:conf/naacl/YasunagaRBLL21,DBLP:conf/iclr/0001BYRLML22}.
A lighter alternative replaces LM scoring with a dot product between GloVe-50 embeddings of the entity and the QA context~\citep{DBLP:conf/nips/YasunagaBR0MLL22}, applying an additional pre-cap of 300 candidates before the final 200-node truncation (\cref{tab:per_system_procedure}).

\begin{table}[h]
\centering
\resizebox{\columnwidth}{!}{
\small
\begin{tabular}{@{}llll@{}}
\toprule
\textbf{System} & \textbf{Scoring} & \textbf{Pre-cap} & \textbf{Final cap} \\
\midrule
QA-GNN   & RoBERTa-large MLM & none & 200 \\
GREASELM & RoBERTa-large MLM & none & 200 \\
DRAGON   & GloVe-50 dot prod. & 300  & 200 \\
\bottomrule
\end{tabular}
}
\caption{\textbf{Subgraph pruning across QA-GNN, GREASELM, and DRAGON.}}
\label{tab:per_system_procedure}
\end{table}

\section{Injection Mechanism Taxonomy}
\label{app:injection_taxonomy}
\Cref{tab:injection_taxonomy} summarizes how the reviewed systems leverage UMLS at the point of model contact. 
The three paradigms are not strictly disjoint in principle --- a system could, for instance, linearize at inference and also encode at training time --- but in the reviewed corpus each system is dominated by a single mechanism, and we classify by the primary route through which UMLS reaches the model.

\begin{table}[t]
\centering
\footnotesize
\setlength{\tabcolsep}{4pt}
\resizebox{0.48\textwidth}{!}{%
\begin{tabular}{llcc}
\toprule
\textbf{Class} & \textbf{Paradigm} & \textbf{How UMLS reaches the model} & \boldmath{$n$} \\
\midrule
I   & Textual linearization & Serialized factual context & 5 \\
II  & Structural injection  & Explicit graph at inference & 9 \\
III & Parameter encoding    & Training-time supervision   & 7 \\
\midrule
    & \textbf{Total}        &                              & \textbf{21} \\
\bottomrule
\end{tabular}
}
\caption{\textbf{Injection mechanism choices.} Counts are restricted to systems with explicit UMLS injection mechanisms.}
\label{tab:injection_taxonomy}
\end{table}

\section{UMLS Evidence Evaluation Setup}
\label{app:downstream_umls_details}

We conduct an LLM-based experimental analysis on MedQA, MedMCQA, and PubMedQA to examine which UMLS construction choices make the retrieved evidence usable for MQA.
The evidence-format comparison spans ten readers from five families---Qwen3.5, Llama~3.1, MedGemma, Gemma~4, and GPT-OSS---covering both general-purpose and medically adapted models.
The path-length comparison is run on the Qwen3.5 collection alone, which varies scale within a single family and therefore isolates scale-dependence from family-level differences.
All results are reported as percentage accuracy gains over the zero-shot baseline.
Accordingly, each intervention is evaluated only on the subset of items that the corresponding zero-shot reader answers incorrectly.
We evaluate the MedQA test split ($n{=}1{,}273$), the MedMCQA development split ($n{=}4{,}183$), and the PubMedQA expert-labeled set ($n{=}500$); MedMCQA test labels are not public, so its development split is used throughout.
This design focuses the analysis on cases where parametric knowledge alone is insufficient, thereby reducing contamination effects and providing a more controlled estimate of the contribution of UMLS evidence.
The prompt used for all evaluations is reported in~\cref{fig:umls_medical_prompt}.

\begin{figure}[h!]
\centering
\small
\begin{tcolorbox}[colframe=black, colback=gray!5,
    title=Medical UMLS Prompt]
\texttt{You are a medical expert tasked with answering multiple-choice questions.} \\[4pt]

\texttt{Use the question stem and, when present, the provided UMLS subgraph as evidence. Select the single best answer and output the answer option letter inside \textbackslash boxed\{\}.}

\texttt{UMLS evidence:} \\
\texttt{\{umls\_subgraph\}} \\[4pt]

\texttt{Question:} \\
\texttt{\{question\}} \\[4pt]

\texttt{Options:} \\
\texttt{\{options\}} \\[4pt]

\texttt{Now answer the question.}
\end{tcolorbox}
\caption{\textbf{Zero-shot prompt with UMLS evidence for medical multiple-choice question answering.}}
\label{fig:umls_medical_prompt}
\end{figure}

\paragraph{Inference Setup}
We evaluate all reader families through a local vLLM endpoint.
Table~\ref{tab:inference_setup} summarizes the inference configuration.

\begin{table}[t]
\centering
\fontsize{8.5}{10}\selectfont
\begin{tabularx}{\columnwidth}{@{}lX@{}}
\toprule
\textbf{Setting} & \textbf{Value} \\
\midrule
Serving & vLLM endpoint with prefix caching and chunked prefill \\
Context length & \texttt{max\_model\_len=65536} \\
Batching & \texttt{max\_num\_seqs=64}, \texttt{max\_num\_batched\_tokens=65536} \\
Cache & FP8 key-value cache \\
Decoding & \texttt{temperature=1.0}, \texttt{top\_p=0.95}, \texttt{top\_k=20}, \texttt{min\_p=0.0}, \texttt{presence\_penalty=1.5}, \texttt{repetition\_penalty=1.0} \\
\bottomrule
\end{tabularx}
\caption{\textbf{Inference configuration.} Serving and decoding settings used for all reader evaluations.}
\label{tab:inference_setup}
\end{table}

\paragraph{Hardware Setup}

Experiments were run on a workstation with Ubuntu 24.04.4 LTS, an Intel Core i9-10900X CPU, 54 GiB RAM, and one NVIDIA RTX PRO 6000 Blackwell GPU with 96 GB of VRAM.
The NVIDIA driver version was 595.58.03 and the CUDA version was 13.2.

\section{Release Sensitivity and Prompting}
\label{app:extended_results}

\paragraph{Isolating the Knowledge Base}

The measurements in \cref{tab:release_source_coverage} require changing the UMLS configuration while holding everything else fixed.
We use scispaCy for this specific comparison because it decouples mention detection from concept linking: the NER model that proposes spans is separate from the knowledge base that resolves them, so the same detected mentions can be linked against a different release or source subset.
Any change in the linked set is then attributable to the UMLS configuration alone.
This isolation is not available in MetaMap, which performs mention extraction and Metathesaurus lookup jointly, so the detected spans themselves depend on the release; MetaMap therefore cannot separate a linker effect from a knowledge-base effect, which is why it anchors the coverage comparison in \cref{sec:nel} rather than this one.
We evaluate the MedQA development split ($n{=}1{,}272$) and the full PubMedQA dataset.
PubMedQA contributes question-side coverage only, since its \textit{yes}/\textit{no}/\textit{maybe} responses correspond to no linkable answer concepts.

\paragraph{Prompting Regime}

The evidence-format comparison in \cref{fig:evidence_form_model_sweep} uses answer-only prompting, in which the reader emits the option letter without an intervening reasoning trace.
Because serialized evidence has to be read before it can be used, the prompting regime is a plausible moderator of the effect, and we probed it with a chain-of-thought pilot on a subsample of the MedQA test split, with model-internal reasoning disabled where the model exposes that control so that the contrast does not conflate prompting with hidden reasoning budgets (\cref{tab:prompting_regime}).
As the winning evidence view changes between regimes, the choice of evidence format cannot be settled independently of how the reader is prompted.
This further finding reinforces the central point that the measured benefit of \quotes{UMLS injection} is a property of a specific construction evaluated under a specific reading procedure, not of the resource.

\paragraph{Path Length}

\cref{fig:path_depth_model_sweep} reports the question-to-answer path-length intervention discussed in \cref{sec:tbu}.
It is run on the Qwen3.5 scale ladder, only on items where two-, three-, and four-hop paths are simultaneously available under the same construction setup, to control for item-level variation. 

\end{document}